\documentclass[lettersize,journal]{IEEEtran}

\usepackage[UKenglish,USenglish]{babel}
\usepackage[T1]{fontenc}
\usepackage[utf8]{luainputenc}

\usepackage[protrusion=true,expansion=true,tracking=true]{microtype}
\usepackage[style=USenglish]{csquotes}
\usepackage{textcomp}

\usepackage{cite}

\usepackage{graphicx}
\graphicspath{{img/}}

\usepackage{amsmath,amssymb,amsfonts}
\newcommand*\rfrac[2]{{}^{#1}\!/_{#2}}

\usepackage{algorithmic}

\usepackage{array}
\usepackage{tabularx}
\usepackage{booktabs}
\usepackage{multirow}
\newcolumntype{Y}{>{\raggedleft\let\newline\\\arraybackslash\hspace{0pt}}X}
\newcolumntype{Z}{>{\centering\let\newline\\\arraybackslash\hspace{0pt}}X}

\usepackage[caption=false]{subfig}

\usepackage{float}

\usepackage{tikz}

\usepackage{url}
\usepackage{hyperref}
\hypersetup{colorlinks=true,linkcolor=blue,citecolor=blue,filecolor=blue,urlcolor=blue}
\usepackage{enumitem}
\setlist[description]{noitemsep}
\setlist[enumerate]{noitemsep}
\setlist[itemize]{noitemsep,topsep=0pt}

\usepackage{xcolor}

\usepackage{lipsum}

\begin{document}

\title{Improving Presentation Attack Detection for ID Cards on Remote Verification Systems}

\author{
    Sebastian Gonzalez,~\IEEEmembership{Member,~IEEE,}
    and Juan Tapia,~\IEEEmembership{Member,~IEEE}\\
    \textbf{This paper is under revision and consideration for a scientific journal}.%
        
\thanks{Juan Tapia, da/sec-Biometrics and Internet Security Research Group, Hochschule Darmstadt, Germany, e-mail: (juan.tapia-farias@h-da.de).}%
\thanks{Sebastian Gonzalez, TOC Biometrics R\&D  Center, Santiago, Chile, email: sebastian.gonzalez@tocbiometrics.com}%
\thanks{Manuscript received xx; revised xx.}}

\markboth{Journal of \LaTeX\ Class Files,~Vol.~14, No.~8, August~2021}%
{Shell \MakeLowercase{\textit{et al.}}: A Sample Article Using IEEEtran.cls for IEEE Journals}


\maketitle

\begin{abstract}
In this paper, an updated two-stage, end-to-end Presentation Attack Detection method for remote biometric verification systems of ID cards, based on MobileNetV2, is presented. Several presentation attack species such as printed, display, composite (based on cropped and spliced areas), plastic (PVC), and synthetic ID card images using different capture sources are used. This proposal was developed using a database consisting of 190.000 real case Chilean ID card images with the support of a third-party company. Also, a new framework called PyPAD, used to estimate multi-class metrics compliant with the ISO/IEC 30107-3 standard was developed, and will be made available for research purposes. Our method is trained on two convolutional neural networks separately, reaching BPCER\textsubscript{100} scores on ID cards attacks of 1.69\% and 2.36\% respectively. The two-stage method using both models together can reach a BPCER\textsubscript{100} score of 0.92\%.
\end{abstract}


\section{Introduction}

\IEEEPARstart{S}{everal} remote biometric verification systems have been developed in the last few years due to the COVID-19 pandemic raising the problem of physically attending stores and banks, social security organisations, and others, in order to request products and/or benefits. Even when attending physically, users would not feel comfortable using fingerprint verification methods that require people to touch the same sensor. 
Due to this problem, remote biometric verification systems based on government-issued documents have been used to remove these limitations, which allow users to compare a selfie image from a smartphone and a face picture extracted from their ID card, passport, driver's license and others. 
The selection of the proper ID document depends directly on each country. For instance, in the USA, the driver's license is broadly adopted for travel and many other operations. In the case of Chile, the driver's license is not used for identity verification by itself, and is only utilised to ascertain that the person in question is authorised to drive a vehicle.
The type of ID document is also a factor in the deployment of products related to the information contained in embedded chips standardised by ICAO regulations\footnote{\url{https://www.icao.int/publications/Documents/9303_p3_cons_en.pdf}}.

Most ID documents in South America are chipless, which causes difficulties in developing systems based on this technology because it is usually either not enabled, or the ID document in question is a printed paper. 
Furthermore, remote biometric verification systems are incrementally targeted by attackers because users are unsupervised, and impostors can try to fool the system numerous times without many limitations in place.

The automated discovery of presentation attacks through Presentation Attack Detection (PAD) mechanisms plays a key role in fighting against malicious actors who aim to interfere with the normal operation of biometric identity verification systems, as the increased adoption of these systems---especially as we mentioned before, those built to work remotely without supervision over the user---are often used to control the access to privileged information (such as bank accounts) and are thus prime targets for biometric fraud and impersonation. PAD algorithms are usually built as classification systems trained to discriminate between bona fide presentations (genuine user transactions) and presentation attacks, where the latter can include a number of potential different spoofing techniques, referred to as Presentation Attack Instrument Species (PAIS) that are dependant on the biometric modality used by the identity verification system. 

PAD algorithms are considered non-zero effort systems because the attacker needs to spend time in developing the type of attack; either spoofing the face using a mask, or a printed/display replay of a selfie; or spoofing the ID card using a modified version of the document. In the later a printed or display replay of the ID card can also be utilised by the attacker.
The principles for the assessment and reporting of the performance of PAD algorithms are defined in the ISO/IEC 30107-3 standard~\cite{iso30107}.

In this work, we present an improved PAD system for Chilean ID cards. In Chile, there are two formats of ID cards in simultaneous operation, one is chipless and does not follow the ICAO standard (hereinafter referred to as \enquote{CHL1}), and the other has an embedded chip and does follow the ICAO standards (hereinafter referred to as \enquote{CHL2}). Therefore, the development of a PAD system capable of generalising and detect attacks for both types of ID cards is relevant and eliminates the effort to keep two systems available.

In order to create a high-quality system, it is also essential to have a toolkit to calculate and report metrics that follows the ISO/IEC 30107-3 standard for PAD system assessment. However, according to our literature review, a specific toolkit focused on the assessment of multi-class PAD systems is not available. For this reason, we present a new Python toolkit and demonstrate an analysis of the results of the latest iteration of our previous work in ID card Presentation Attack Detection~\cite{gonzalez2020hybrid}, using a Convolutional Neural Network (CNN) method for the classification of bona fide, printed, display, composite, plastic, and synthetic images of Chilean ID cards.

The main contributions of this work can be summarised as follows:

\begin{itemize}
    \item \emph{Two-stage network}: A serial, two-stage architecture trained from scratch is proposed. This means no ImageNet~\cite{imagenet_cvpr09} weights were used in this proposal. This architecture consists of one MobileNetV2 for detecting between bona fide, composite, and synthetic ID cards, and a second CNN used to detect bona fide versus print, display, and plastic PAIS. A concatenation of both models is proposed as a final application.

    \item \emph{End-to-end system}: This paper reports an end-to-end system. Which includes the image capture system, ID card segmentation system, and PAD method. This is a significant difference from the state-of-the-art works where many assumptions are included.
    
    \item \emph{Class weights}: Balanced class weights were used to correctly represent the quantity of presentation attacks instrument samples per species in order to overcome the class imbalance present in the dataset.

    \item \emph{Database}: This paper presents results using a real remote biometric system database. We increased the number of bona fide Chilean ID cards and presentation attack instruments used from previous approaches from 27,000 up to approximately 190,000 images. Also two new species of presentation attacks were added: Plastic and Synthetic.

    \item \emph{Data augmentation (DA)}: An aggressive DA technique was used to train both CNNs from scratch. These images allow the network to learn more challenging scenarios, including different lighting conditions, blurring, gaussian noise, coarse occlusion, cropped images, and others.

    \item \emph{PAD assessment toolkit}: A new toolkit for the assessment of multi-class PAD systems is presented. This new toolkit, called PyPAD, is compliant with ISO 30107-3 requirements and will be available for research purposes (upon acceptance).
\end{itemize}

The rest of the article is organised as follows: Section~\ref{sec:relate} summarises the related works on ID card PAD. Our proposed method for detecting presentation attack instruments of Chilean ID card images is described in Section~\ref{sec:method}. 
The database is described in Section~\ref{sec:database}. The experimental framework and results of this work are then presented in Section~\ref{sec:experiments}. We conclude the article in Section~\ref{sec:conclusion}.

\section{Related work}
\label{sec:relate}

Some works in the literature have tackled the problem of detecting ID card presentation attacks. For instance, Shi and Jain~\cite{shi2018docface, shi2019docface+} proposed DocFace and DocFace+ to determine the authenticity of a personal ID document by comparison of a face image (selfie) with the photo ID in the ID document. In DocFace+, a kiosk scans the ID document photos or reads the photo from the embedded chip using an NFC reader.

Zheng et al.~\cite{Zheng2019ASO} present a survey and provide an overview on typical image tampering methods, released image tampering datasets, and recent tampering detection approaches. It presents a distinct perspective to rethink various assumptions about tampering clues, which can be discovered by different detection approaches. This further encourages the research community to develop general tampering localisation methods in the future instead of adhering to single-type tampering detection. Most of the analysis was realised using handcrafted images in different domains. Deep learning methods have not been explored in detail.

Albiero et al.~\cite{albiero} present a method for comparing selfie images to photo ID images from Chilean ID cards across adolescence, employing fine-tuning techniques using a private dataset.

Bulan et al.~\cite{Bulan} propose a cryptography-based system for authenticating the content in printed images, where an encrypted thumbnail of the image is generated and embedded in the printed image, which can be later extracted and decrypted from a scan of the printed image.

Stokke et al.~\cite{stokkenes2018biometric} proposed an online banking authentication system based on features extracted from faces using bloom filters. This information is encoded and used as a key for accessing banking services.

Perera et al.~\cite{perera2019face} proposed an active authentication system that monitors user identity after initial access is granted continuously. A similar approach has recently been reported by Fathy et al.~\cite{fathy2015face}.

Arlazarov et al.~\cite{MIDV-500} presented a tiny ID card dataset containing 500 video clips of 50 different identity document types. This dataset was one of the first made publicly available for identity document analysis and recognition in the video stream. Additionally, the paper presents three experimental baselines obtained using the dataset: face detection accuracy, separate text fields OCR precision for four major identity document field types, and identity document data extraction from video clips.

Zhu et al.~\cite{zhu} propose a method that indicates whether the photo ID or data fields in the ID card image have been altered or replaced by digital or handcrafted means. On the other hand, if the source of the ID card comes from a printing/scanning process or the image was captured from a digital screen, it means that it does not come from the original plastic document, and alterations could have been made beforehand.

Mudgalgundurao et al.~\cite{DBLP:journals/iet-bmt/RajaR022} proposed a method to detect fake German ID cards and residence permits using pixel-wise supervision based on DenseNet. This technique enables the method to leverage minute cues on various artefacts, such as moiré patterns and artefacts left by the printers. The authors present the baseline benchmark using different handcrafted and deep learning models on a newly constructed in-house database obtained from an operational system consisting of 886 users with 433 bona fide, 67 print and 366 display attacks. This work presents several assumptions regarding image input before it is classified.

In our previous approach, Gonzalez et al.~\cite{gonzalez2020hybrid} proposed a hybrid two-stage classification system that checks if the entire Chilean ID card was tampered with or modified. In that work, the authors analysed the primary sources of fraud as image composition and image source tampering based on ImageNet weights~\cite{imagenet_cvpr09}. The paper is relevant because it presents results on genuine transactions of a remote verification system with 24,778 images distributed across different species including bona fide, composite, printed, and display ID card images.
This new approach by far outperforms our previous work in Gonzalez et al.~\cite{gonzalez2020hybrid}. 

This new proposal includes an image auto-capture stage and a new segmentation stage that removes the background from the ID card picture captured by each user in the remote operation. Also, we increase the number of images from 27,000 to 190,000. New attacks, such as synthetics and plastic (PVC) ID cards, were also added. On the other hand, a new PyPAD library was created to evaluate the ID card PAD system. Two independent convolutional neural networks were trained separately, and then the final evaluation was performed by concatenating both models. This two-stage approach allows us to improve the results to the point where the operating threshold of the system can be set according to the BPCER\textsubscript{100} score instead of the BPCER\textsubscript{10} or BPCER\textsubscript{20} scores.

\subsection{Evaluation metrics}
\label{sec:eval}

Several implementations can be found in the literature to measure PAD systems, such as BOSARIS~\cite{brummer2013bosaris}, SIDEKIT ~\cite{larcher2016extensible} and PyEER\footnote{\url{https://pypi.org/project/pyeer/}}. These toolkits have been used to report metrics such a the confusion matrix, Area Under the Curve (AUC), Detection Trade-off curves (DET), Equal Error Rate (EER), Bona Fide Presentation Classification Error Rate (BPCER), and Attack Presentation Classification Error Rate (APCER). However, they are not created specifically for PAD on ID cards.

The BOSARIS Toolkit provides MATLAB code for calibrating, fusing and evaluating scores from (automatic) binary classifiers~\cite{brummer2013bosaris}. It was developed to provide solutions for automatic speaker recognition. Still, they envision much of the code will have wider applicability for other biometric and/or forensics problems, where the calibration of likelihood ratios is of interest.

SIDEKIT is an open-source Python toolkit that includes a large assembly of state-of-the-art components and allows fast prototyping of an end-to-end speaker recognition system~\cite{larcher2016extensible}. Each step includes front-end feature extraction, normalisation, speech activity detection, modelling, scoring and visualisation. This tool offers a wide range of standard algorithms and flexible interfaces. The use of a single efficient programming and scripting language (Python in this case), and the limited dependencies, facilitate the deployment of industrial applications and extensions to include new algorithms.

PyEER is a python package intended for performance evaluation of biometric systems oriented for fingerprint application, but it can also be used to evaluate binary classification systems. It was developed to provide researchers and the scientific community with a general tool to correctly evaluate and report their systems' performance.

In order to improve and complement these proposals, we present a new multi-class Python toolkit called PyPAD. This new toolkit will be available by request for research purposes only (upon acceptance). The new features are described in section~\ref{sec:eval}.

\section{Database}
\label{sec:database}

The database consists of images of two Chilean ID card formats available, which we refer to as \enquote{CHL1} and \enquote{CHL2}, comprising bona fide, printed, displayed (on smartphones, tablets, and monitor screens), composite, plastic (PVC card), and synthetic images. Bona fide ID card images were acquired from voluntary users of a fully remote biometric enrolling and identity verification system; this corresponds to real users in a real-world scenario and covers a high range of illumination variability. 

This system utilises its own image auto-capture process in order to standardise image resolution to $1280\times720$ pixels with no post-processing, regardless of the device employed, as official camera apps integrated on the device often use median/beautification filters that result in an undesired loss of pixel detail and feature information.

For the source system, the printed ID card images were created by printing bona fide images on either plain or high-glossy paper. Images were scaled to the size of real Chilean ID cards when printing, resulting in eight images per paper, which were then cut manually and captured in different illumination scenarios using the same capture process utilised by the remote identity verification system. Plastic PVC ID card images were obtained by using an ID card printer, previously selecting most frontal (less skewed) bona fide images and segmenting them (see Section~\ref{subsec:segmentation}).

For the display images, a diverse set of smartphone, tablet and monitor screens were used to display bona fide ID card images, which were then captured one at a time using the same capture process previously mentioned.

Composite ID cards for border system correspond to images where the identity of the original owner of the ID card is changed, either by swapping over the authentic photo ID in the document with the photo ID of another subject or by altering the written information contained in the original ID card through image processing methods. 

Face detection on ID card images was done automatically by employing an MTCNN model~\cite{zhang2016joint} to extract the photo ID coordinates for the splicing process. 

For the text, EasyOCR\footnote{\url{https://github.com/JaidedAI/EasyOCR}} was used to extract three regions of the document: the RUN (unique identification number given to Chilean residents); the region containing first and last names; and the region containing nationality, sex, birth date, document issue date, document ID (only for the CHL2 format), and document expiration date. The composition method spliced either the photo ID or the three data regions of one document and pasted it into another. The transition was either unprocessed, meaning that the photo ID or data regions of the origin ID card were only scaled to the size of the target ID card and pasted, leaving images with noticeable borders around the composited areas, or processed using OpenCV’s seamless cloning\footnote{\url{https://docs.opencv.org/4.6.0/df/da0/group__photo__clone.html}} method to smooth the transition between the photo ID or the regions during composition.

Synthetic ID cards were generated from a high-quality Chilean ID card template, which was manually modified to remove the photo ID and text containing the subject's personal information. A program was developed to select a random face image based on \cite{danben_synt} and then paste it over the photo ID section of the document. The same program also fills the subject data fields with a random combination of information from different lists, imitating the text style of an original Chilean ID card. This process is run multiple times to create different synthetic ID card images.

Additionally, a number of tampered ID card images recovered from actual impostor attacks on the remote identity verification system were added. These attack presentation instruments were not created by our team but rather by real antagonists. 

Table~\ref{tab:database} shows the total count of images for both ID card formats and all species in the database. Unfortunately, this is a sequestered database and will not be available for commercial use, but an evaluation of a third-party proposal can be performed using the test set.

\begin{table*}[!htb]\def\tabularxcolumn#1{m{#1}}
    \centering
	\caption{Chilean ID card images database.}
	\label{tab:database}
	\begin{tabularx}{\linewidth}{rrYYYY}
    	\toprule
        \multirow{2}{*}{\textbf{Document type}} & \multirow{2}{*}{\textbf{Class}} & \multicolumn{3}{c}{Split} & \multirow{2}{*}{\textbf{Total per class}} \\
        \cmidrule(lr){3-5}
        && \textbf{Train} & \textbf{Validation} & \textbf{Test} & \\
    	\midrule
    	\multirow{6}{*}{\textsc{CHL1}} & \textbf{Bona fide} &  9,526 & 3,250 & 3,170 & 15,946 \\
    	& \textbf{Print}                                    &  9,774 & 3,252 & 3,280 & 16,306 \\
    	& \textbf{Display}                                  & 12,124 & 4,026 & 4,050 & 20,200 \\
    	& \textbf{Composite}                                &  7,779 & 2,592 & 2,980 & 13,351 \\
    	& \textbf{Plastic}                                  &  1,768 &   589 &   590 &  2,947 \\
    	& \textbf{Synthetic}                                &  9,931 & 3,310 & 3,310 & 16,551 \\
    	\cmidrule(lr){2-6}
    	\multirow{6}{*}{\textsc{CHL2}} & \textbf{Bona fide} & 11,613 & 3,831 & 3,840 & 19,284 \\
    	& \textbf{Print}                                    & 12,341 & 4,053 & 4,090 & 20,484 \\
    	& \textbf{Display}                                  & 13,299 & 4,393 & 4,440 & 22,132 \\
    	& \textbf{Composite}                                & 13,669 & 4,532 & 4,250 & 22,451 \\
    	& \textbf{Plastic}                                  &  1,878 &   590 &   590 &  3,058 \\
    	& \textbf{Synthetic}                                &  9,931 & 3,310 & 3,310 & 16,551 \\
    	\midrule
    	& \textbf{Total per split} & 113,633 & 37,728 & 37,900 & \textbf{\textsc{189,261}} \\
    	\bottomrule
	\end{tabularx}
\end{table*}

\begin{figure*}[!htb]
    \centering
    \subfloat[CHL1 Bona fide]{\includegraphics[width=0.25\textwidth]{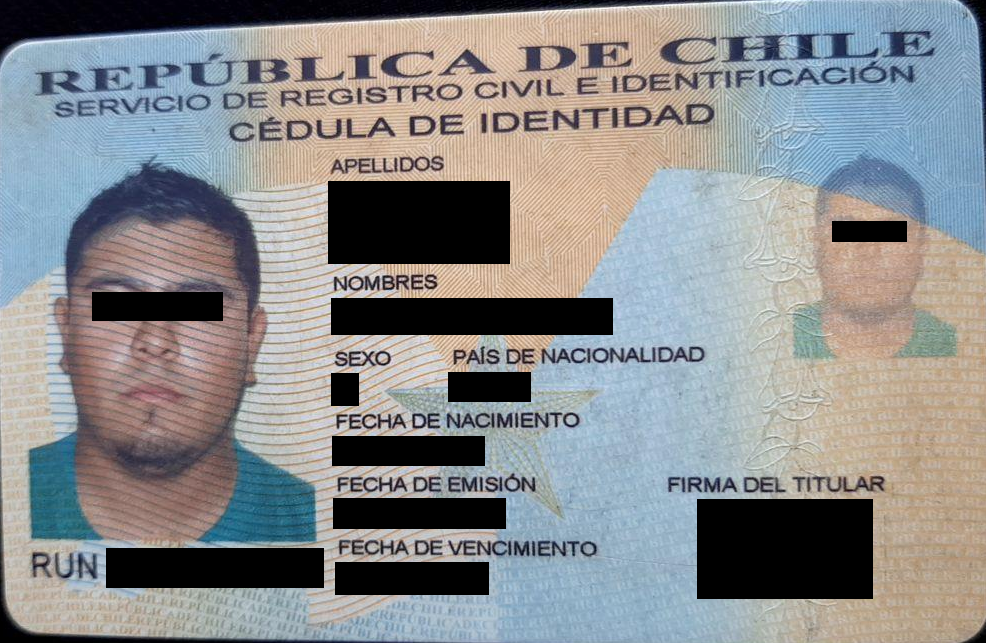}}\hfil%
    \subfloat[CHL1 Print]{\includegraphics[width=0.25\textwidth]{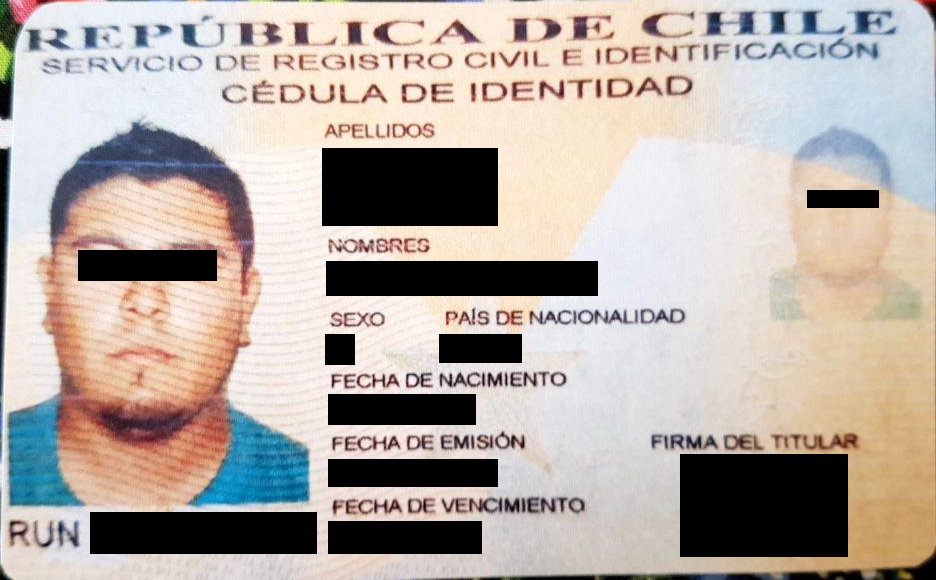}}\hfil%
    \subfloat[CHL1 Display]{\includegraphics[width=0.25\textwidth]{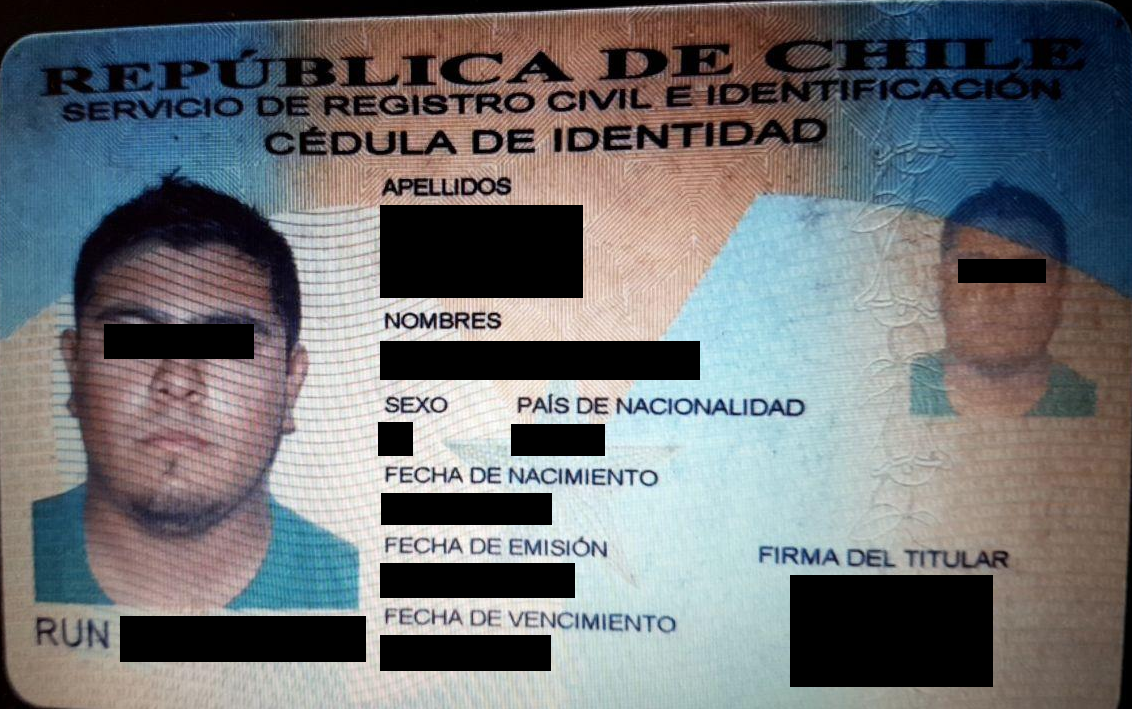}}\\
    \subfloat[CHL1 Composite]{\includegraphics[width=0.25\textwidth]{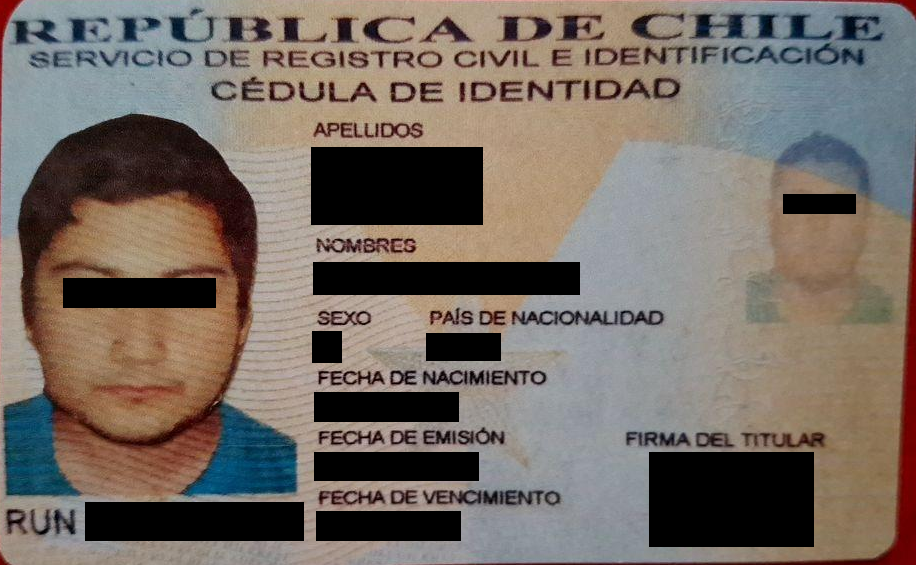}}\hfil%
    \subfloat[CHL1 Plastic]{\includegraphics[width=0.25\textwidth]{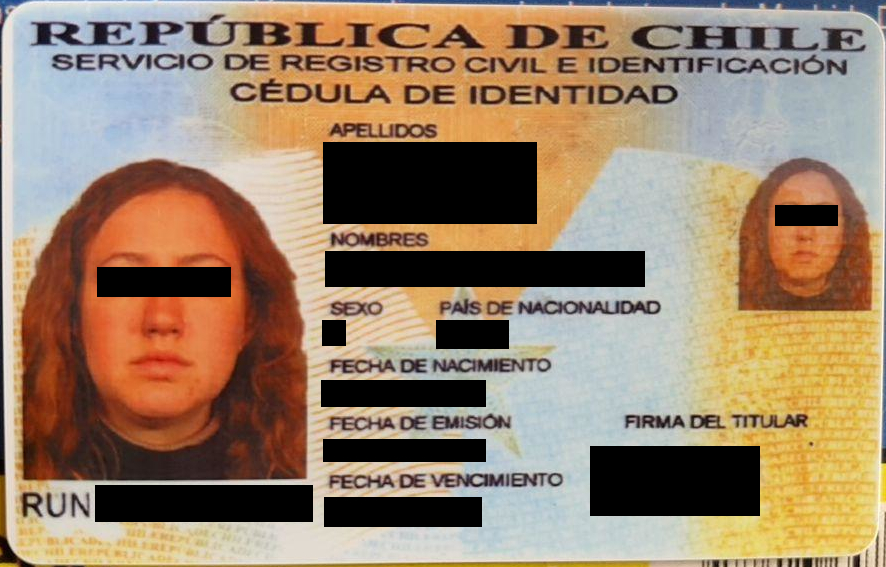}}\hfil%
    \subfloat[CHL1 Synthetic]{\includegraphics[width=0.25\textwidth]{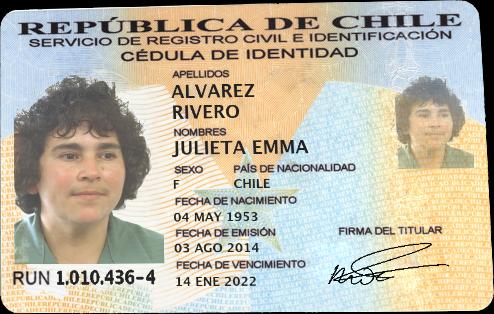}}%
    \caption{Example of Chilean ID cards. (a) to (f) showcase examples of bona fide, print, display, composite, plastic, and synthetic images belonging to the \enquote{CHL1} ID card format. Sensitive data has been blacked out, with the exception of synthetic cards, which contain fabricated data.}
    \label{fig:specieschl1}
\end{figure*}

\begin{figure*}[!htb]
    \centering
    \subfloat[CHL2 Bona fide]{\includegraphics[width=0.25\textwidth]{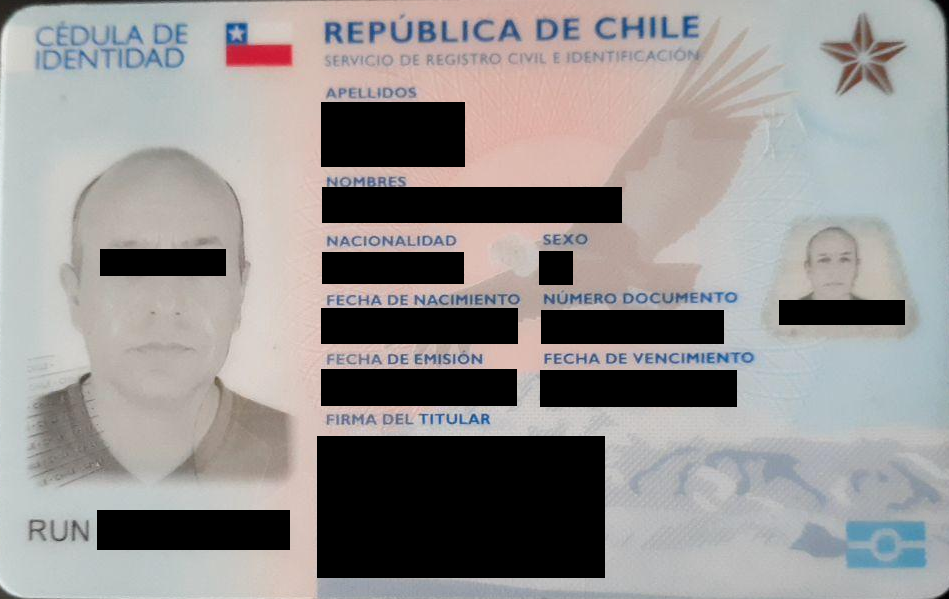}}\hfil%
    \subfloat[CHL2 Print]{\includegraphics[width=0.25\textwidth]{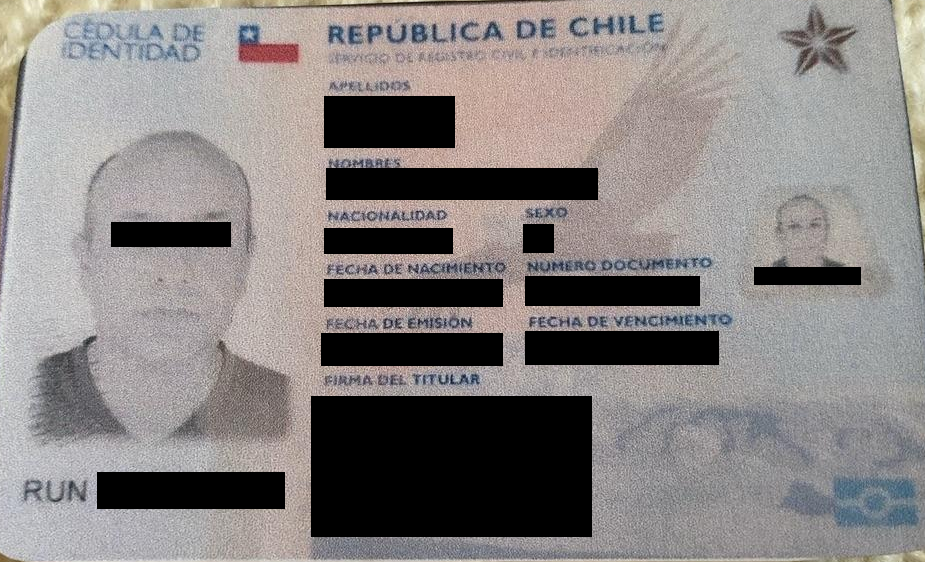}}\hfil%
    \subfloat[CHL2 Display]{\includegraphics[width=0.25\textwidth]{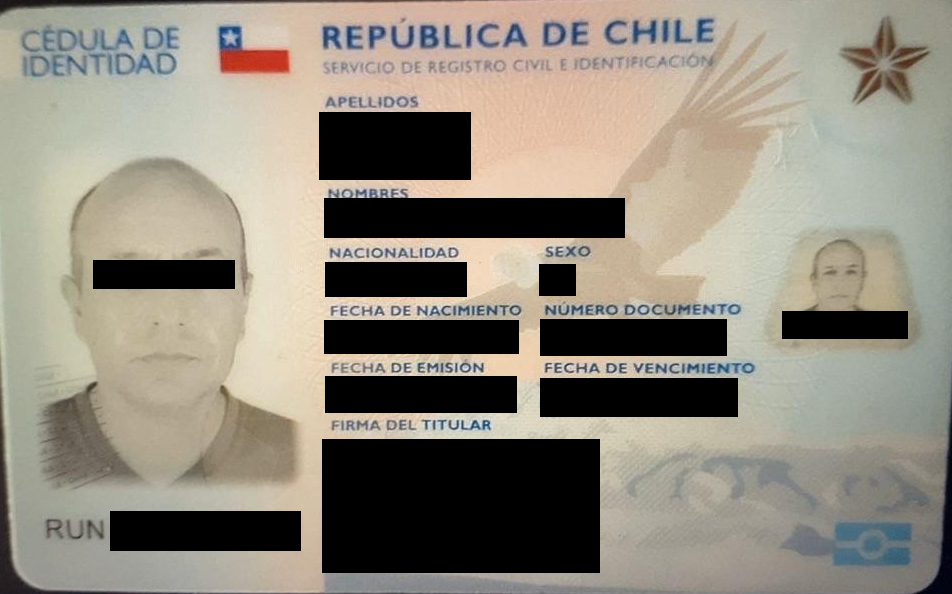}}\\%
    \subfloat[CHL2 Composite]{\includegraphics[width=0.25\textwidth]{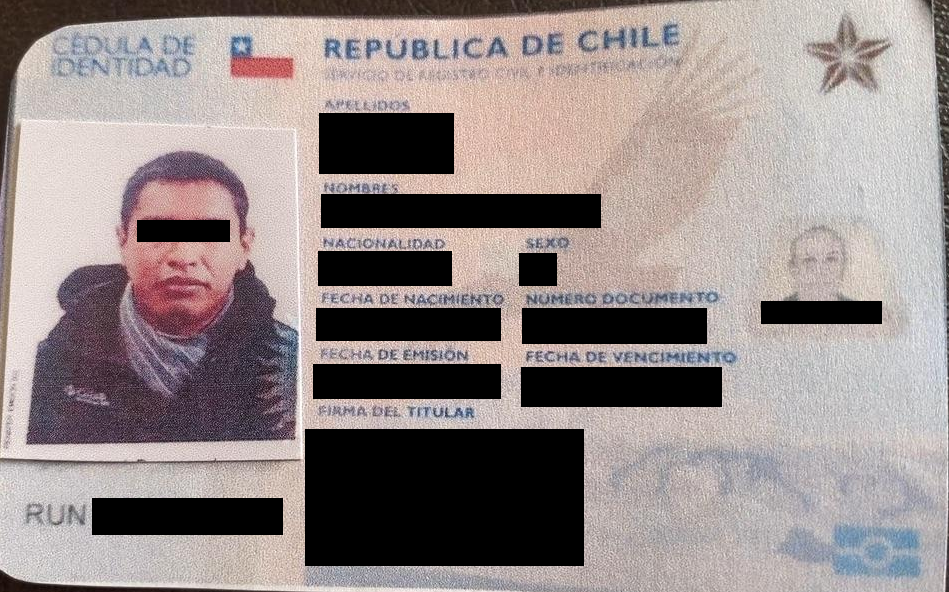}}\hfil%
    \subfloat[CHL2 Plastic]{\includegraphics[width=0.25\textwidth]{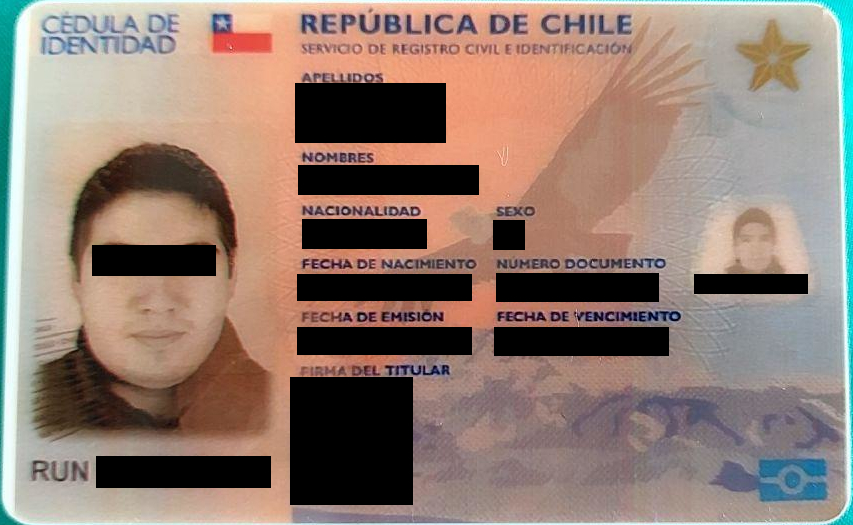}}\hfil%
    \subfloat[CHL2 Synthetic]{\includegraphics[width=0.25\textwidth]{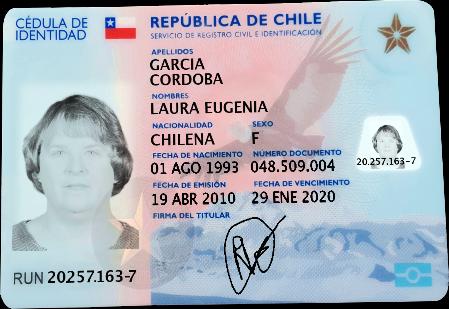}}%
    \caption{Example of Chilean ID cards. (a) to (f) showcase examples of bona fide, print, display, composite, plastic, and synthetic images belonging to the \enquote{CHL2} ID card format. Sensitive data has been blacked out, with the exception of synthetic cards, which contain fabricated data.}
    \label{fig:specieschl2}
\end{figure*}

Figures~\ref{fig:specieschl1} and~\ref{fig:specieschl2} illustrate a single example image per species for each ID card format (CHL1 and CHL2). Please note that sensitive information has been redacted for purposes of illustration.

\section{Method}
\label{sec:method}

This paper proposes a two-stage Deep Learning network for PAD of spoofed Chilean ID cards. This system was built to train two classifiers independently, one focusing on composite and synthetic attacks, and another focusing on printed and display attacks. Each classifier was trained and evaluated separately using segmented images. In the end, we concatenate both classifiers in order to build a strong and robust PAD system.
MobileNetV2 was selected as a classifier based on lightweight and fast deployment for real-time applications. Also, this selection makes it feasible to compare this work with the preliminary approach presented by~\cite{gonzalez2020hybrid}.

\subsection{MobileNetV2}

The MobileNetv2 netwrok was trained from scratch using segmented images, in order to remove the background and only retain pixels belonging to the ID cards in the image. Each image was resized after that to the size of $448\times448\times3$ pixels. A different number of filters were explored in order to look for simple and complex artefacts in the images. Fine-tuning analyses were made but obtained sub-optimal performance. A learning rate of $1\times 10^5$, Adam optimiser, and categorical cross-entropy were used with a batch size of 64. The optimal values were selected based on a grid search.
All the models in the experiments for this paper have been trained in a server with two GTX 2080-Ti GPUs with 64 GB RAM.

\subsection{Image segmentation}
\label{subsec:segmentation}

\begin{figure}[!htb]
    \centering
    \includegraphics[width=0.35\textwidth]{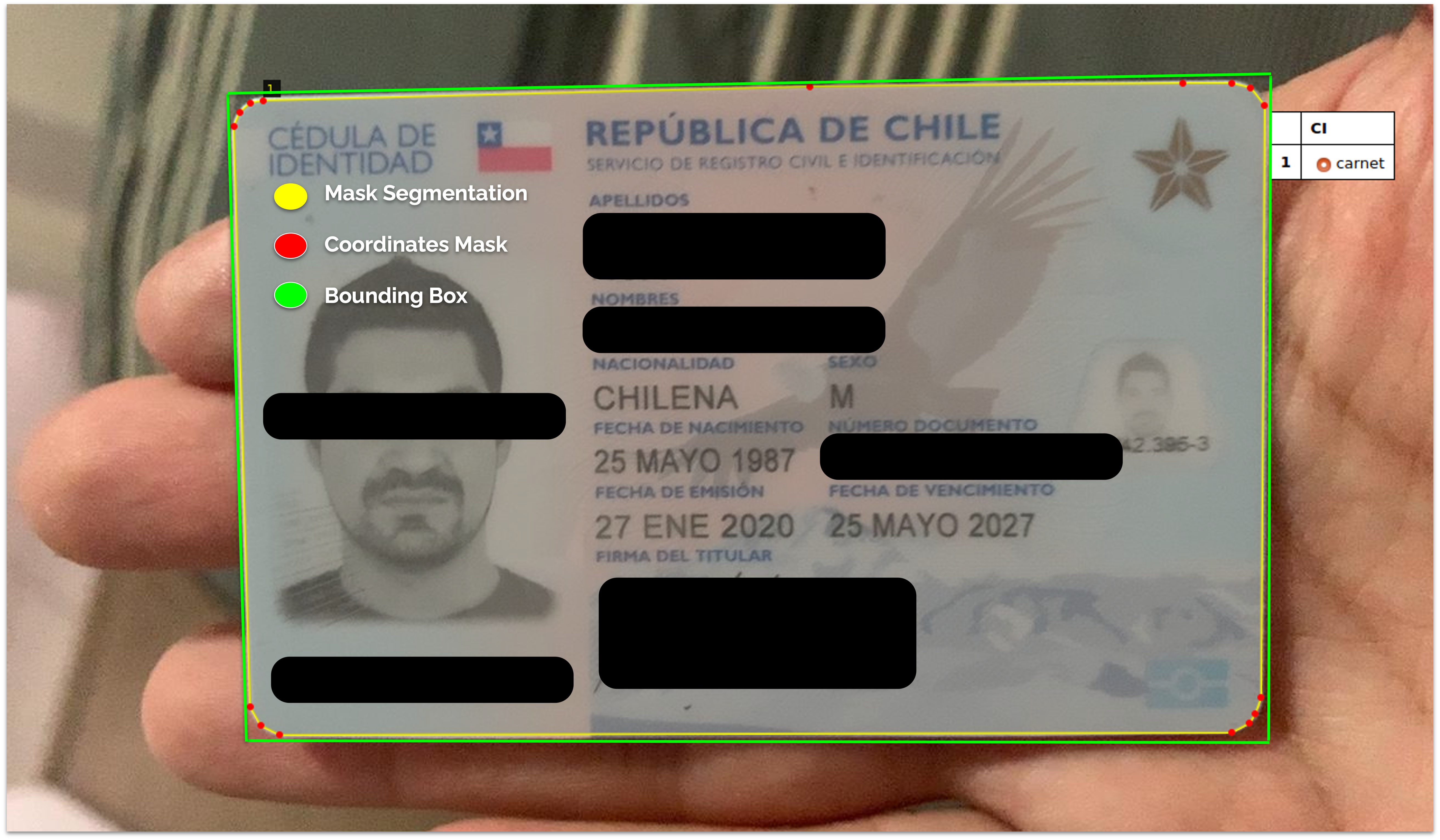}
    \caption{Example of segmented ID card with manual and automatic annotations.}
    \label{fig:segm}
\end{figure}

Semantic segmentation was applied to the ID card images in order to remove the background pixels when training and evaluating the networks. This process is carried out with the aim of making the network focus only on the pixels belonging to the ID card in the image, which helps improve classification and attack detection. The image semantic segmentation process is achieved based on a lightweight MobileUNet~\cite{ronneberger2015unet} CNN trained specifically with RGB ID card images from several countries, including Chile, Argentina, and Mexico. The boundaries of the ID card in each image were manually annotated for each sample. Afterwards---by using the manual annotations---the background is removed, and another one previously selected is applied to the images, creating challenging background variations for each ID card image. See Figure~\ref{fig:segm}.

Furthermore, in order to approach the variability in lighting and colour temperature present in the ID card images acquired from the identity verification system previously mentioned, alterations in hue and saturation are performed for each image. This is accomplished by changing the colour space to HSV and choosing an angle between -10° to 18° for the H component and a multiplier between 0.9 and 1.18 for the S component. Finally, a colour space conversion back to RGB is performed. A detailed explanation of the segmentation algorithm can be found in~\cite{lara2021towards}. 

\subsection{Data Augmentation}

An aggressive Data Augmentation (DA) method was applied when training the modified MobileNetV2 networks. A large number of  operations, such as changes in colours and lighting conditions, affine and perspective transformations, contrast changes, gaussian noise, random dropout of image regions, cropping/padding, and blurring, were included in the training dataset to represent the real conditions for a remote verification system. These DA operations are based on the imgaug library\footnote{\url{https://github.com/aleju/imgaug}}, which is optimised for high performance. This improves the quality of the training results by using very challenging images.

\subsection{Class Weights}

Regularly, an algorithm may present bias towards the majority class (the one with the most samples in the training data) if the dataset is unbalanced, thus deteriorating the performance of the model. In order to overcome the class imbalance present in the dataset, class weights are computed according to the number of samples of each class, where samples of the most prevalent classes will have weights inversely proportional to the occurrence of the class.

Effectively, class weights are applied to the loss function by re-scaling the gradient steps during training, favouring under-represented classes with fewer samples while penalising over-represented classes with more samples. See Equation~\ref{eqn:weights}.

\begin{equation}
\label{eqn:weights}
    Weight_i = \frac{Nsamples}{Nclasses \times samples_i}
\end{equation}

Where $Weight_i$ is the weight for class $i$, $Nsamples$ is the total number of images in the database, $Nclasses$ is the total number of classes in the database, and $samples_i$ is the number of samples of class $i$. The class weights used for the first network the following: Composite: 0.9705, Bona fide: 0.9847, Synthetic: 1.0480. The class weights for the second network are the following: Bona fide: 1.1404, Print/Plastic: 0.9358, Display: 0.9483.

\subsection{Metrics}
\label{sec:metrics}

The ISO/IEC 30107-3 standard~\cite{iso30107} details the methodology for PAD algorithm performance assessment for biometric systems through various evaluation metrics, which have been implemented in PyPAD. Bona Fide Classification Error Rate (BPCER) measures the proportion of bona fide presentations incorrectly classified as attack presentations by the biometric system. BPCER is computed according to Equation~\ref{eq:bpcer}, where $N_{BF}$ corresponds to the total number of bona fide presentation images, and $RES_{i}$ for the $i$th image is one if the system classifies it as an attack presentation, or 0 if it is classified as a bona fide presentation~\cite{marcel2019handbook}. This is subject to the operational point $\tau$ of the system, where a classification score for the bona fide class greater than the decision threshold will result in the sample being classified as a bona fide presentation, whereas a score less than or equal to $\tau$ will result in the sample being determined as a presentation attack by the system.

\begin{equation}\label{eq:bpcer}
    BPCER(\tau) = \frac{\sum_{i=1}^{N_{BF}}RES_{i}}{N_{BF}}
\end{equation}

The attack Presentation Classification Error Rate (APCER) measures the proportion of attack presentations erroneously classified as bona fide presentations. This metric is calculated for each PAIS, and only the worst-case scenario is considered. Equation~\ref{eq:apcer} details how to compute the APCER, in which the value of $N_{PAIS}$ corresponds to the number of attack presentation images for a given PAIS. This metric is also subject to the operational point $\tau$ of the system.

\begin{equation}\label{eq:apcer}
    APCER_{PAIS}(\tau) = \frac{1}{N_{PAIS}} \sum_{i=1}^{N_{PAIS}} (1-RES_{i})
\end{equation}

The Equal Error Rate (EER) represents the trade-off when the APCER\textsubscript{PAIS}($\tau$) is equal to the BPCER($\tau$) at some operational point $\tau$. However, an operational point $\tau$ resulting in equal BPCER and APCER may not exist for the system: in this case, an interpolated decision threshold can be used. A threshold that minimises the distance between both rates is also acceptable.

These metrics effectively measure to what degree the PAD algorithm confuses attack presentations with bona fide presentations and vice versa. Furthermore, the Average Classification Error Rate (ACER) yields the average between the APCER and the BPCER. This is shown in Equation~\ref{eq:acer}. The ACER is typically used to evaluate the overall performance of the biometric system, but it has been deprecated in the ISO/IEC 30107-3 and is thus not compliant with the standard.

\begin{equation}\label{eq:acer}
    ACER(\tau) = \frac{APCER(\tau)+BPCER(\tau)}{2}
\end{equation}

The selection of an operational point $\tau$ for the biometric system is not simple and heavily depends on the system's use case. For this reason, it is common to report the BPCER\textsubscript{AP} (where AP refers to the attack potential) for PAD algorithms instead, which corresponds to the BPCER when the APCER is fixed to a value of $\rfrac{100}{AP}$. For example, BPCER\textsubscript{10} corresponds to the BPCER when the APCER is fixed at 10\%, and BPCER\textsubscript{20} which is the BPCER when the APCER is fixed at 5\%. This is specially useful given that the APCER of the system can be set at a fixed rate in which the proportion of attack presentations incorrectly classified as bona fide presentations is acceptable, instead of choosing an operational point $\tau$. For example, a remote biometric verification system used for bank account enrolment will want to keep the APCER at a very strict low rate, since this kind of system can be a desirable target for potential attackers, and failing to reject impostors can have serious repercussions. The BPCER\textsubscript{AP} allows measuring the biometric system in terms of false rejections of genuine users while fixing the rate of false acceptance of impostors, which is a straightforward manner of choosing an operational threshold $\tau$ for the system. Nonetheless, as is the case with the EER, an operational point $\tau$ resulting in the desired APCER may not exist, in which case either the closest decision threshold or an interpolated threshold can be utilised.


Detection Error Trade-off (DET) curves, commonly used in speaker recognition tasks~\cite{brummer2013bosaris}, are also conventional in PAD assessment and are used as a means to represent the performance of the PAD algorithm as a graphical plot for easy visual assessment, comparing security vs. convenience measures. DET curves feature the APCER in the $X$ axis and the BPCER on the $Y$ axis, and differs from ROC curves by warping the axes using the quantile function of the normal distribution (probit function); the advantage of this transformation is that it emphasises the region of the plot where the critical differences occur. Furthermore, the axes of the plot are generally restricted to the [0, 50] range, since systems with classification error rates higher than 50\% are ineffective, in addition to the fact that the probit function maps probabilities from $[0, 1]$ to $[-\infty, +\infty]$, so some kind of limitation is needed. Examples are presented in Section~\ref{sec:experiments}.

In order to complement other toolkits explained in Section~\ref{sec:eval}, created for general purpose and binary classification, we propose the new toolkit called PyPAD. This toolkit can be utilised to assess the performance of PAD systems using ISO metrics for either binary or multi-class classification. This implementation will be available for other researchers by request (upon acceptance). This toolkit can be downloaded from~\footnote{\url{https://github.com/jedota/PyPAD}}.
The PyPAD toolkit has the following features:

\begin{itemize}
    \item It is fully compliant with ISO 30107-3 and configurable to choose and estimate results based on different thresholds.
    \item It is able to calculate metrics for binary and multi-class PAD systems.
    \item It is able to plot DET curves containing all the presentation attack species for comparison, depicting the two operational points typically reported: BPCER\textsubscript{10} and BPCER\textsubscript{20}, with values highlighted in the plot.    
    \item An EER plot is created automatically that can help to understand the relation between BPCER and APCER and system thresholds.
    \item Kernel Distribution Estimation (KDE) plots are reported using a linear and a log scale to highlight details and thresholds.
    \item Configurable confusion matrices for multi-class problems and different thresholds.
    \item A summary report is automatically generated describing different operational points (BPCER, APCER).
\end{itemize}

\section{Experiments and results}
\label{sec:experiments}

Three experiments were set up to analyse the results of our proposal and show the suitability of our PyPAD toolkit.
Experiment 1 estimates the results of a three-class PAD problem, classifying bona fide, composite, and synthetic attack species images. The best results can be seen in Table~\ref{tab:border}.

Experiment 2 estimates the results of a four-class PAD problem, classifying bona fide versus print, display, and plastic attack images. The best results can be seen in Table~\ref{tab:source}.

Experiment 3 estimates the results of a five-class PAD problem, classifying one image through both classifiers. In this case a bona fide or attack image is analysed through each network individually to look for modifications. This approach minimises bona fide errors.

For experiments 1 and 2, the operational point $\tau$ is set to the threshold corresponding to the BPCER\textsubscript{100} when presenting the APCER for the non-worse case attack species and the ACER (which is computed using the worse case attack species) of the system.
The ACER is computed mainly for the purpose of comparing with other state-of-the-art works. For experiment 3, the operational point $\tau$ is set to the threshold corresponding to the BPCER\textsubscript{100} for each model separately.

\subsection{Experiment 1 - Composite/Synthetic Detector}

\begin{figure*}[!htb]
    \centering
    \subfloat[Full confusion matrix]{\includegraphics[width=0.3\textwidth]{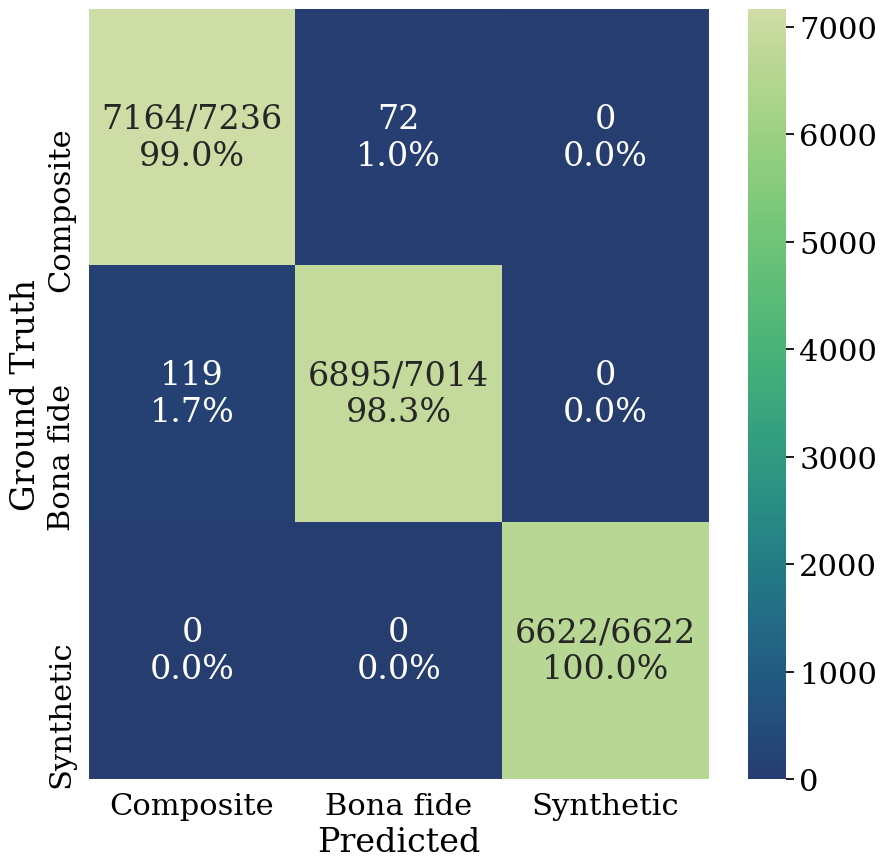}\label{fig:border1a}}\hfil%
    \subfloat[Binary confusion matrix]{\includegraphics[width=0.3\textwidth]{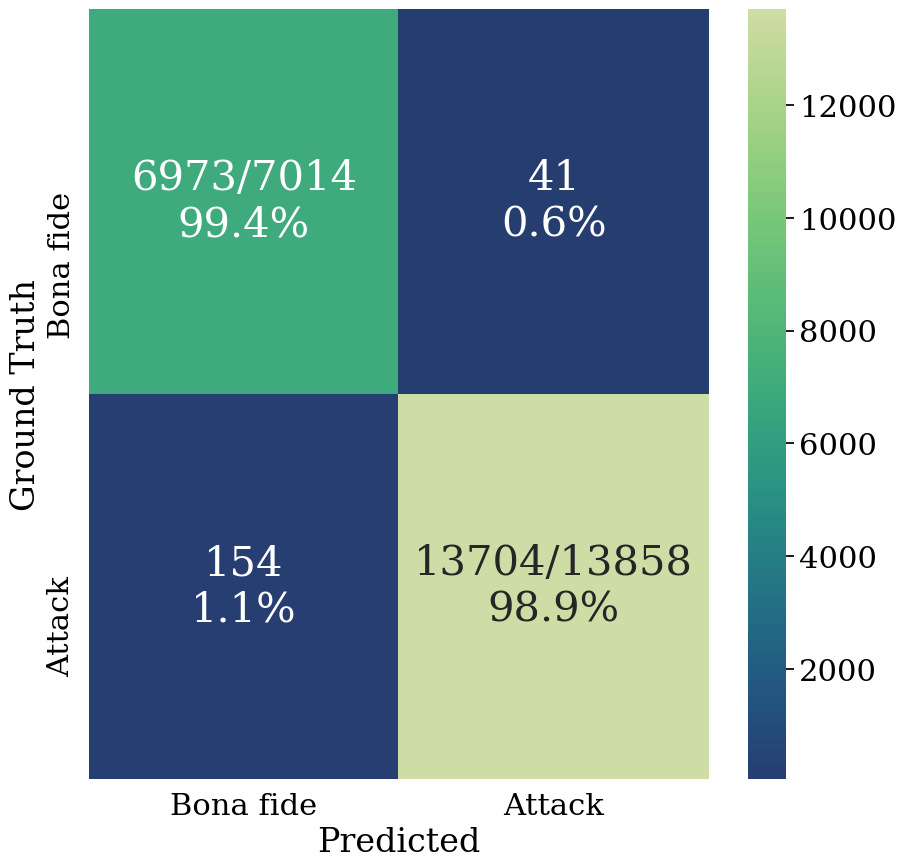}\label{fig:border1b}}\hfil%
    \subfloat[DET curve]{\includegraphics[width=0.3\textwidth]{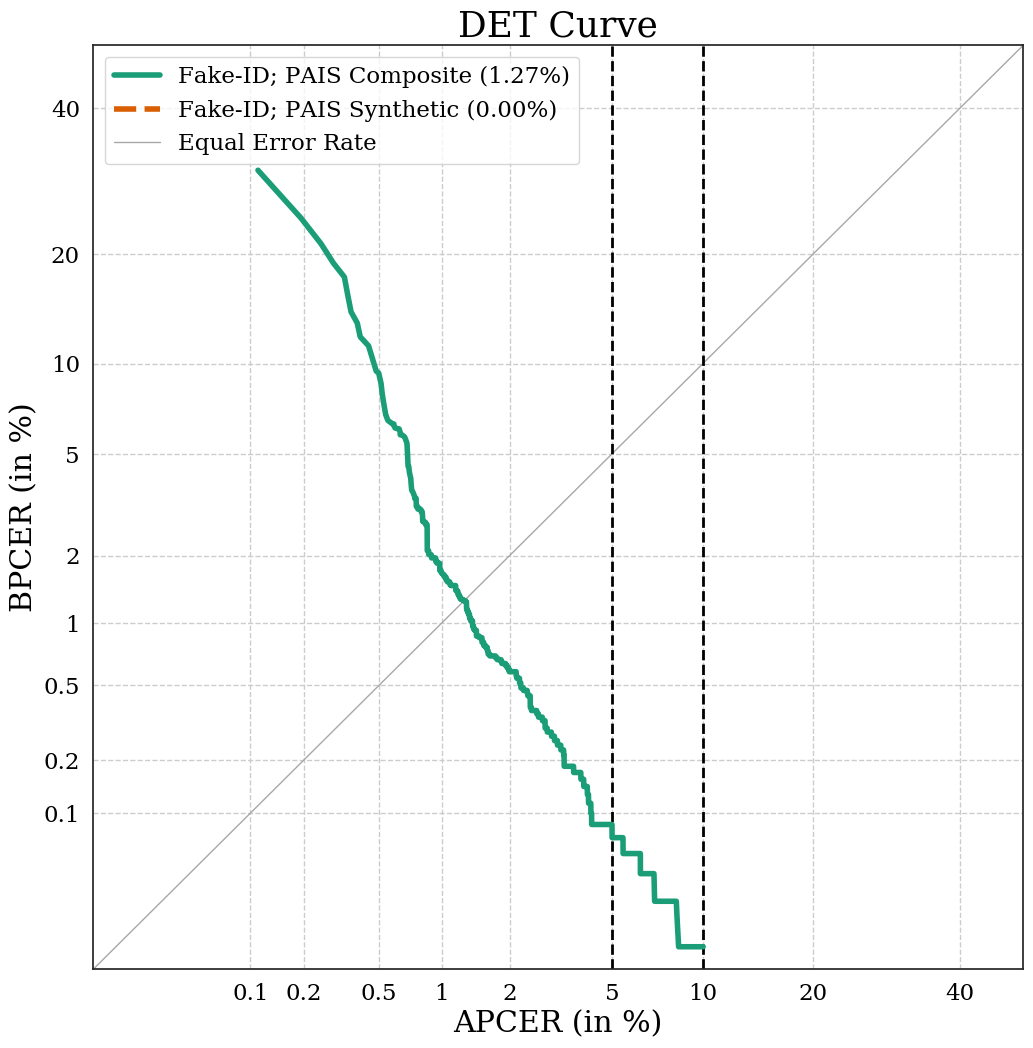}\label{fig:border1c}}\\
    \caption{Border confusion matrices considering each PAI species independently. The second confusion matrix (b) considers the bona fide class versus the fusion of all PAI species. The detection error trade-off curve is depicted in (c). The number in parenthesis corresponds to the EER in percentage. Similarly, The black dashed lines indicate two operational points for BPCER\textsubscript{10} and BPCER\textsubscript{20}.}
    \label{fig:border1}
\end{figure*}

Figure~\ref{fig:border1} shows the results reached for the Composite/Synthetic detector system as a three-class problem (Figure~\ref{fig:border1a}), and a binary problem (Figure~\ref{fig:border1b}), which considers composite/synthetic as one attack class.
Figure~\ref{fig:border1c}, displays the DET curve, identifying the composite species as the worse case scenario, with an EER of 1.27\%. The synthetic images were detected perfectly by the system with an EER of 0.0\%. 

\begin{figure*}[!htb]
    \centering
    \subfloat[Score probability distributions (linear scale)]{\includegraphics[width=0.32\textwidth]{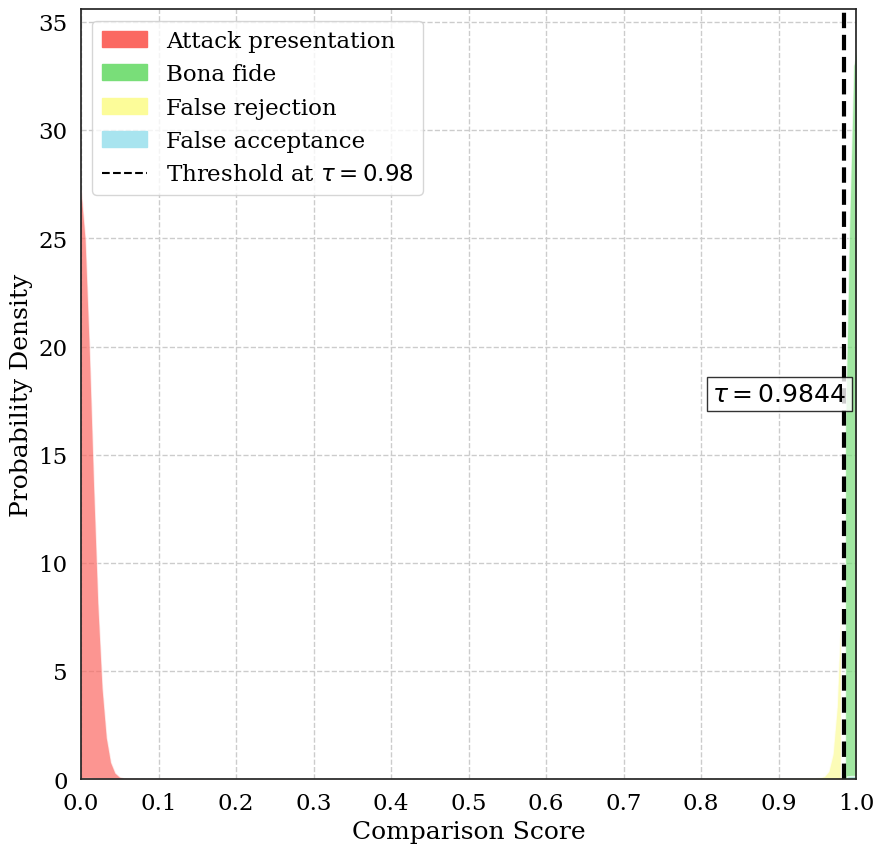}\label{fig:border2a}}\hfil%
    \subfloat[Score probability distributions (log scale)]{\includegraphics[width=0.32\textwidth]{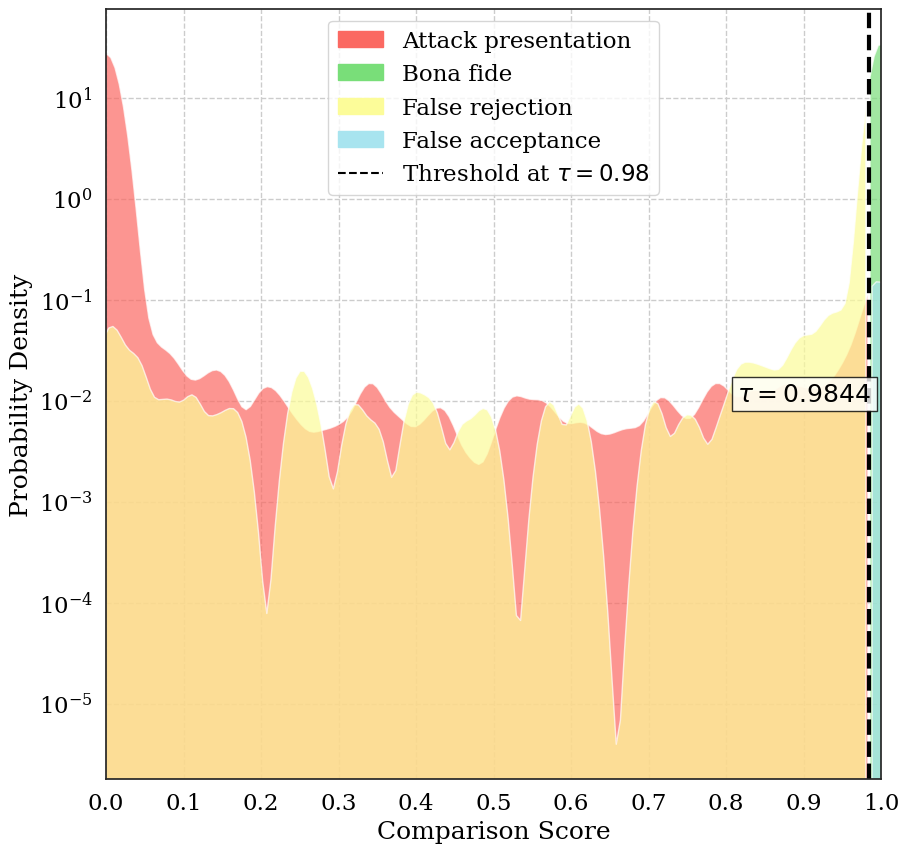}\label{fig:border2b}}\hfil%
    \subfloat[EER curve]{\includegraphics[width=0.32\textwidth]{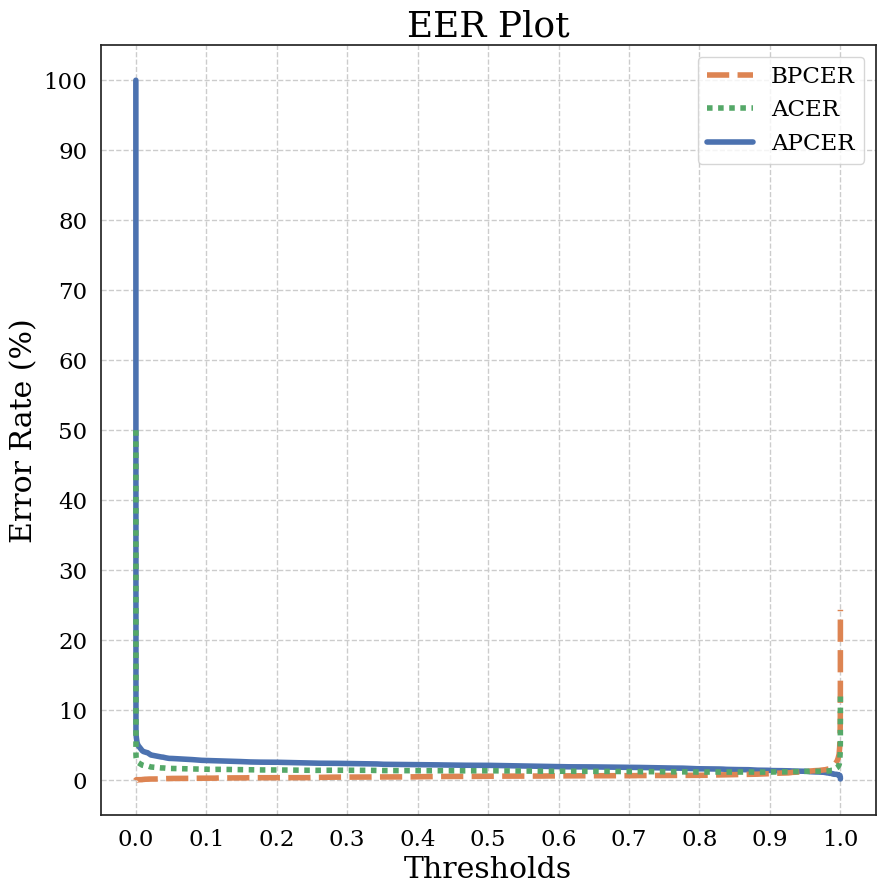}\label{fig:border2c}}\\
    \caption{Border score probability distributions (a and b) and error rates curves (c).}
    \label{fig:border2}
\end{figure*}

Figure~\ref{fig:border2} shows the score probability distributions. Figure~\ref{fig:border2a} shows a KDE in a linear scale, whereas Figure~\ref{fig:border2b} shows a KDE with a log scale on the $Y$ axis used for highlighting minute details. Figure~\ref{fig:border2c} shows APCER and BPCER scores in relation to their respective thresholds in order to visualise the point where the scores converge (EER point). All the figures were created automatically by our PyPAD framework.

Table~\ref{tab:border} shows several operational points, from BPCER\textsubscript{10}, BPCER\textsubscript{20}, BPCER\textsubscript{50}, BPCER\textsubscript{100}, BPCER\textsubscript{500}, BPCER\textsubscript{1000}, and BPCER\textsubscript{10000}. These results can be used to analyse the performance of the classifier network. The corresponding thresholds and EER are also reported. A suitable operational point may be selected according to the application requirements for each use case. The best result selected for the Composite/Synthetic detector was BPCER\textsubscript{100} with a score of 1.68\%.


\begin{table}[!htb]\def\tabularxcolumn#1{m{#1}}
    \centering
	\caption{Experiment 1 - Border CNN PAD assessment.}
	\label{tab:border}
	\begin{tabularx}{\linewidth}{Xrr}
    	\toprule
    	\textbf{Metric} & \textbf{Score (\%)} & \textbf{Threshold} ($\tau$) \\
    	\midrule
    	EER\textsubscript{Composite} &  1.2689 & 0.9523 \\
    	\midrule
    	BPCER\textsubscript{10} (APCER\textsubscript{Composite} = 10\%)          &  0.0143 & 0.0000 \\
    	BPCER\textsubscript{20} (APCER\textsubscript{Composite} = 5\%)           &  0.0855 & 0.0030 \\
    	BPCER\textsubscript{50} (APCER\textsubscript{Composite} = 2\%)           &  0.5845 & 0.5649 \\
    	\textbf{BPCER\textsubscript{100} (APCER\textsubscript{Composite} = 1\%)} & \textbf{1.6823} & \textbf{0.9844} \\
    	BPCER\textsubscript{200} (APCER\textsubscript{Composite} = 0.5\%)        &  9.3532 & 0.9999 \\
    	BPCER\textsubscript{500} (APCER\textsubscript{Composite} = 0.2\%)        & 24.3468 & 1.0000 \\
    	BPCER\textsubscript{1000} (APCER\textsubscript{Composite} = 0.1\%)       & 30.7458 & 1.0000 \\
    	BPCER\textsubscript{10000} (APCER\textsubscript{Composite} = 0.01\%)     & 30.7458 & 1.0000 \\
    	\midrule
    	APCER\textsubscript{Composite}($\tau$) & 0.9950 & 0.9844 \\
    	APCER\textsubscript{Synthetic}($\tau$) & 0.0000 & 0.9844 \\
    	BPCER($\tau$)                          & 1.6823 & 0.9844 \\
    	ACER($\tau$)                           & 1.3387 & 0.9844 \\
    	\bottomrule
	\end{tabularx}
\end{table}

\subsection{Experiment 2 - Print/Plastic/Display Detector}

\begin{figure*}[!htb]
    \centering
    \subfloat[Full confusion matrix]{\includegraphics[width=0.32\textwidth]{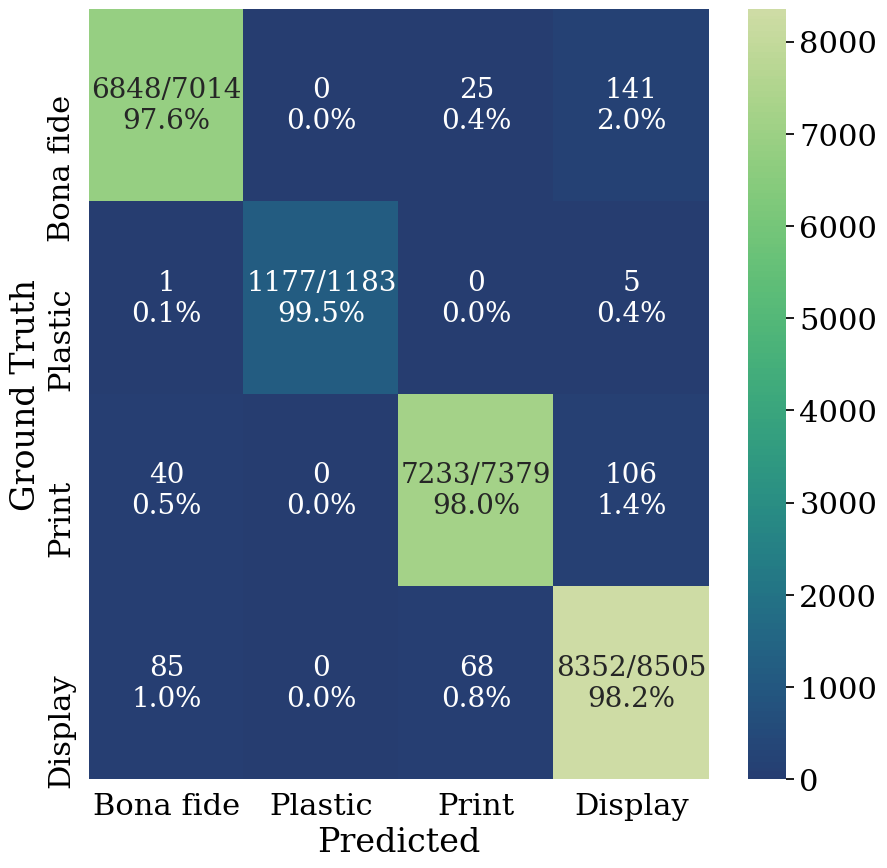}\label{fig:source1a}}\hfil%
    \subfloat[Binary confusion matrix]{\includegraphics[width=0.32\textwidth]{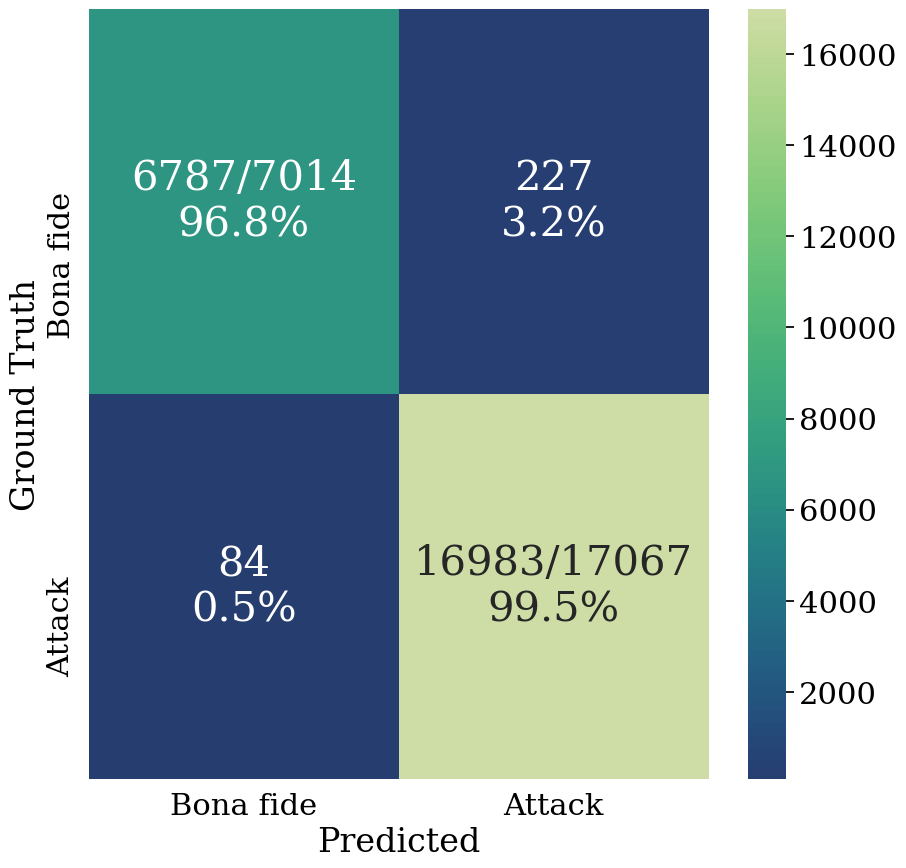}\label{fig:source1b}}\hfil%
    \subfloat[DET curve]{\includegraphics[width=0.32\textwidth]{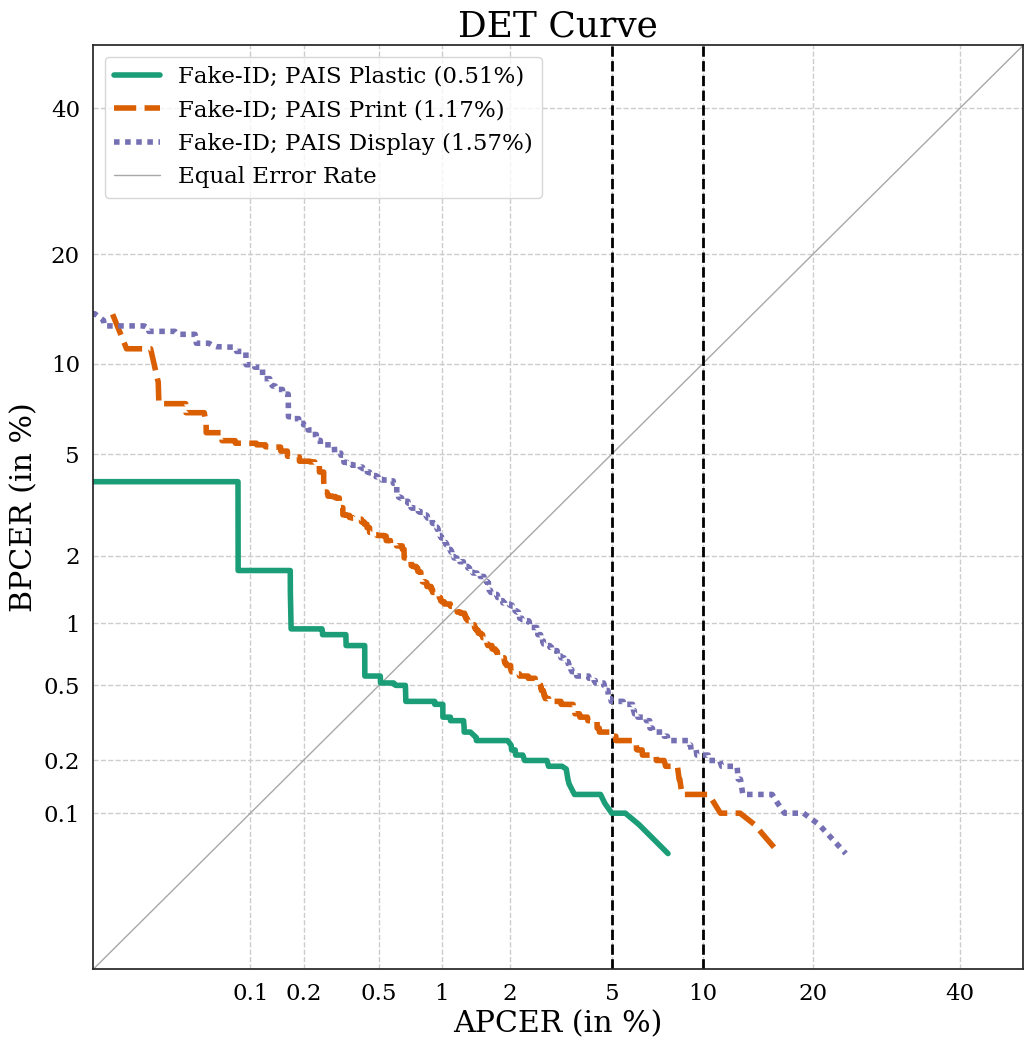}\label{fig:source1c}}\\
    \caption{Source confusion matrices (a and b) considering each PAI species independently. The second confusion matrix (b) considers the bona fide class versus the fusion of all PAI species. The detection error trade-off curve is depicted in (c). The number in parenthesis corresponds to the EER in percentage. Similarly, The black dashed lines indicate two operational points for BPCER\textsubscript{10} and BPCER\textsubscript{20}.}
    \label{fig:source1}
\end{figure*}

Figure~\ref{fig:source1} shows the results reached for the Print/Plastic/Display detector system as a four-class problem (Figure~\ref{fig:source1a}), and a binary problem (Figure~\ref{fig:source1b}), which considers all three attack species as one single class.
Figure~\ref{fig:source1c}, displays the DET curve, identifying the Display species as the worse case scenario, with an EER of 1.57\%. The Print and Plastic species images were detected with an EER of 1.17\% and 0.51\%, respectively.

\begin{figure*}[!htb]
    \centering
    \subfloat[Score probability distributions (linear scale)]{\includegraphics[width=0.3\textwidth]{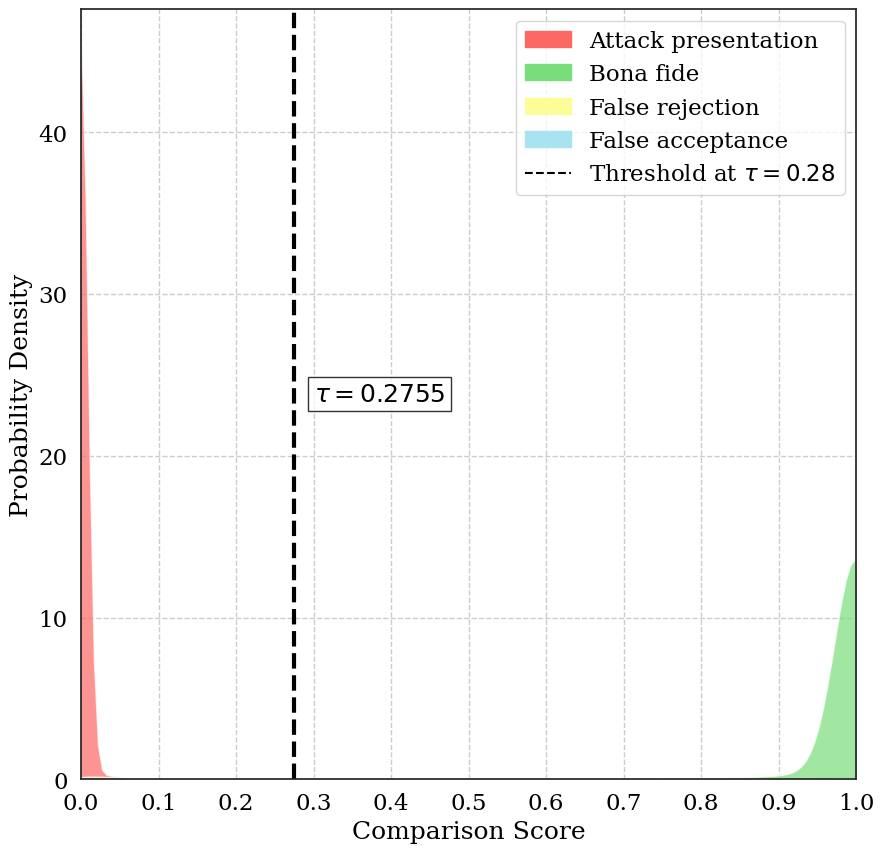}\label{fig:source2a}}\hfil%
    \subfloat[Score probability distributions (log scale)]{\includegraphics[width=0.3\textwidth]{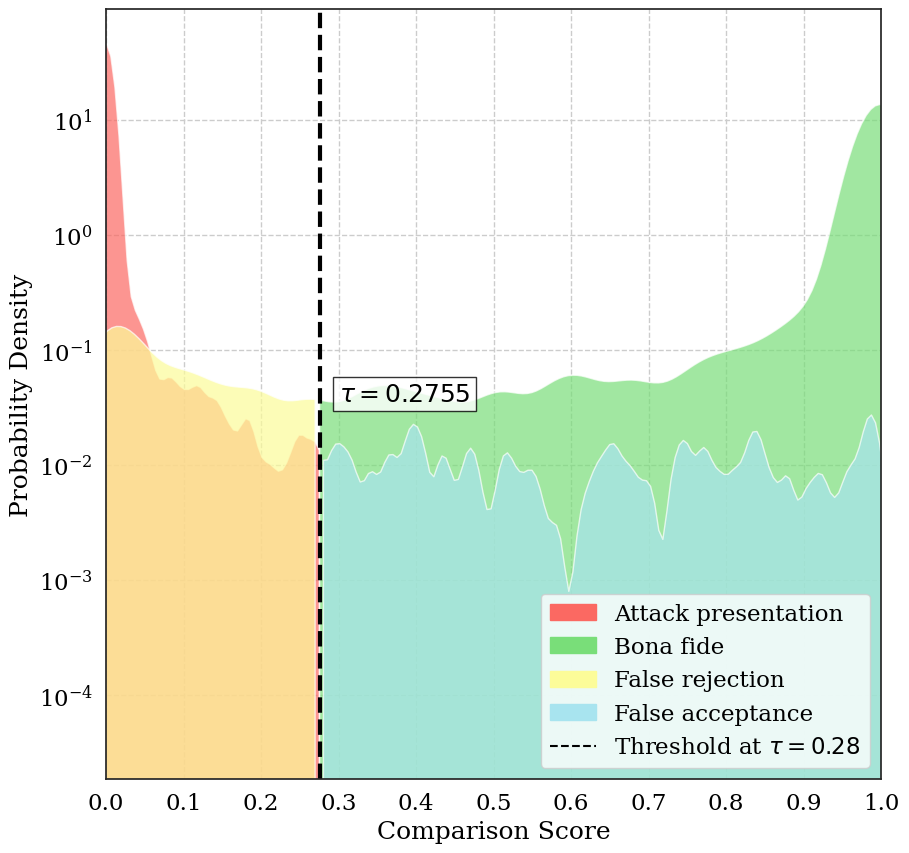}\label{fig:source2b}}\hfil%
    \subfloat[EER curve]{\includegraphics[width=0.3\textwidth]{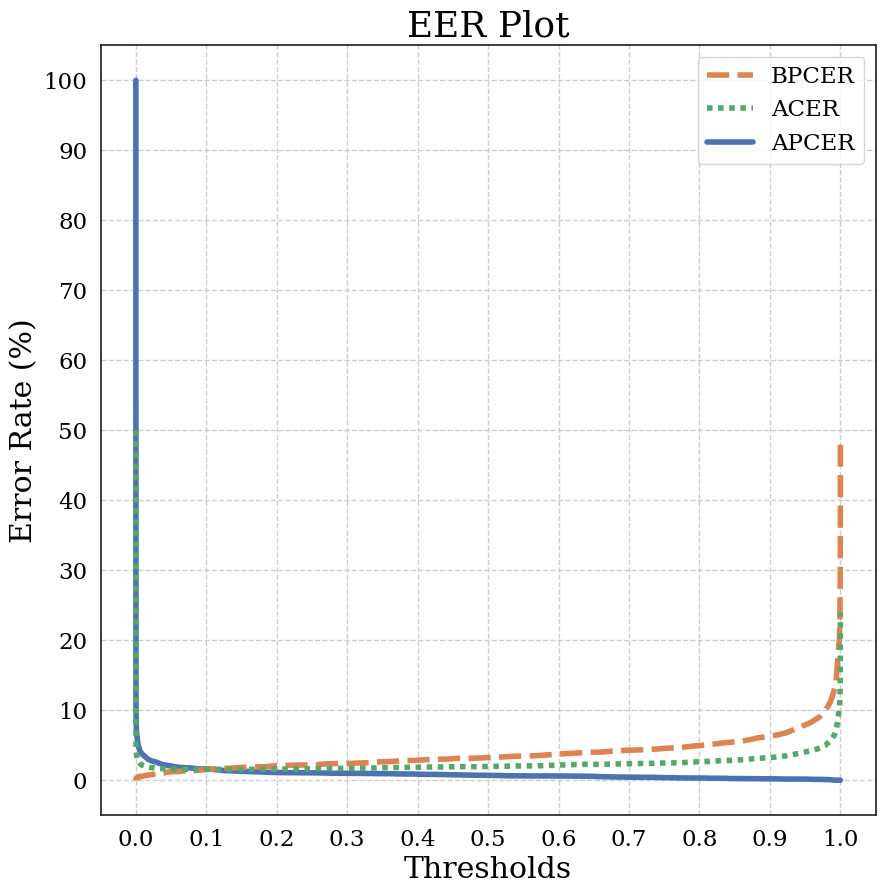}\label{fig:source2c}}\\
    \caption{Source score probability distributions (a and b) and error rates curves (c).}
    \label{fig:source2}
\end{figure*}

Figure~\ref{fig:source2} shows the score probability distributions. Figure~\ref{fig:source2a} shows a KDE in a linear scale, whereas Figure~\ref{fig:border2b} shows a KDE with a log scale. Figure~\ref{fig:source2c} shows APCER and BPCER scores in relation to their respective thresholds in order to visualise the EER point where the scores converge. All the figures were created automatically by our PyPAD framework.

Table~\ref{tab:source} shows several operational points given by BPCER\textsubscript{AP}. The corresponding thresholds and EER are also reported. The best result selected for the Print/Plastic/Display detector was BPCER\textsubscript{100} with a score of 2.36\%.

\begin{table}[!htb]\def\tabularxcolumn#1{m{#1}}
    \centering
	\caption{Experiment 2 - Source CNN PAD assessment.}
	\label{tab:source}
	\begin{tabularx}{\linewidth}{Xrr}
    	\toprule
    	\textbf{Metric} & \textbf{Score (\%)} & \textbf{Threshold} ($\tau$) \\
    	\midrule
    	EER\textsubscript{Display} & 1.5683 & 0.1054 \\
    	\midrule
    	BPCER\textsubscript{10} (APCER\textsubscript{Display} = 10\%)          &  0.2139 & 0.0002 \\
    	BPCER\textsubscript{20} (APCER\textsubscript{Display} = 5\%)           &  0.4135 & 0.0033 \\
    	BPCER\textsubscript{50} (APCER\textsubscript{Display} = 2\%)           &  1.2119 & 0.0506 \\
    	\textbf{BPCER\textsubscript{100} (APCER\textsubscript{Display} = 1\%)} & \textbf{2.3667} & \textbf{0.2755} \\
    	BPCER\textsubscript{200} (APCER\textsubscript{Display} = 0.5\%)        &  4.0776 & 0.6657 \\
    	BPCER\textsubscript{500} (APCER\textsubscript{Display} = 0.2\%)        &  6.3302 & 0.9022 \\
    	BPCER\textsubscript{1000} (APCER\textsubscript{Display} = 0.1\%)       &  9.9230 & 0.9785 \\
    	BPCER\textsubscript{10000} (APCER\textsubscript{Display} = 0.01\%)     & 13.9863 & 0.9937 \\
    	\midrule
    	APCER\textsubscript{Plastic}($\tau$) & 0.0845 & 0.2755 \\
    	APCER\textsubscript{Print}($\tau$)   & 0.5421 & 0.2755 \\
    	APCER\textsubscript{Display}($\tau$) & 0.9994 & 0.2755 \\
    	BPCER($\tau$)                        & 2.3667 & 0.2755 \\
    	ACER($\tau$)                         & 1.6831 & 0.2755 \\
    	\bottomrule
	\end{tabularx}
\end{table}

\subsection{Experiment 3 - Two-stage system}

\begin{figure*}[!htb]
    \centering
    \subfloat[Full confusion matrix]{\includegraphics[width=0.32\textwidth]{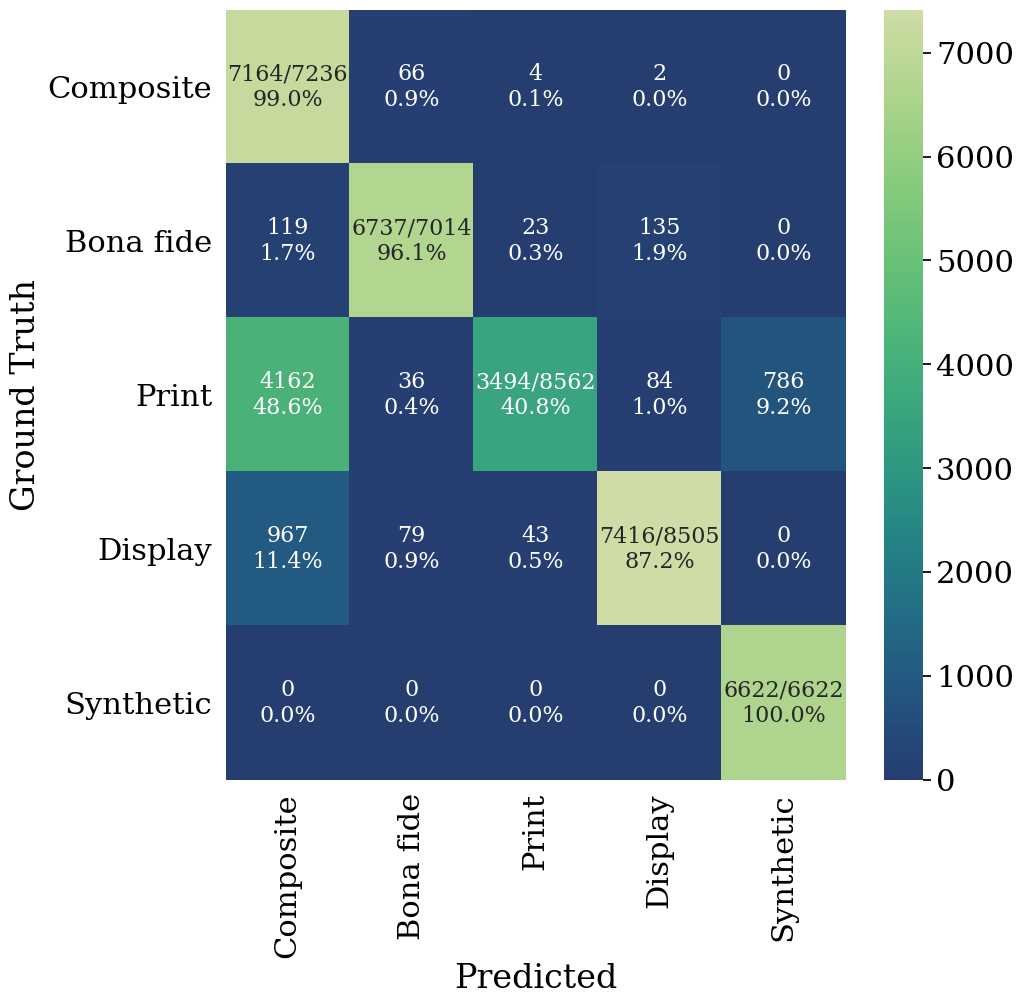}\label{fig:cascade1a}}\hfil%
    \subfloat[Binary confusion matrix]{\includegraphics[width=0.32\textwidth]{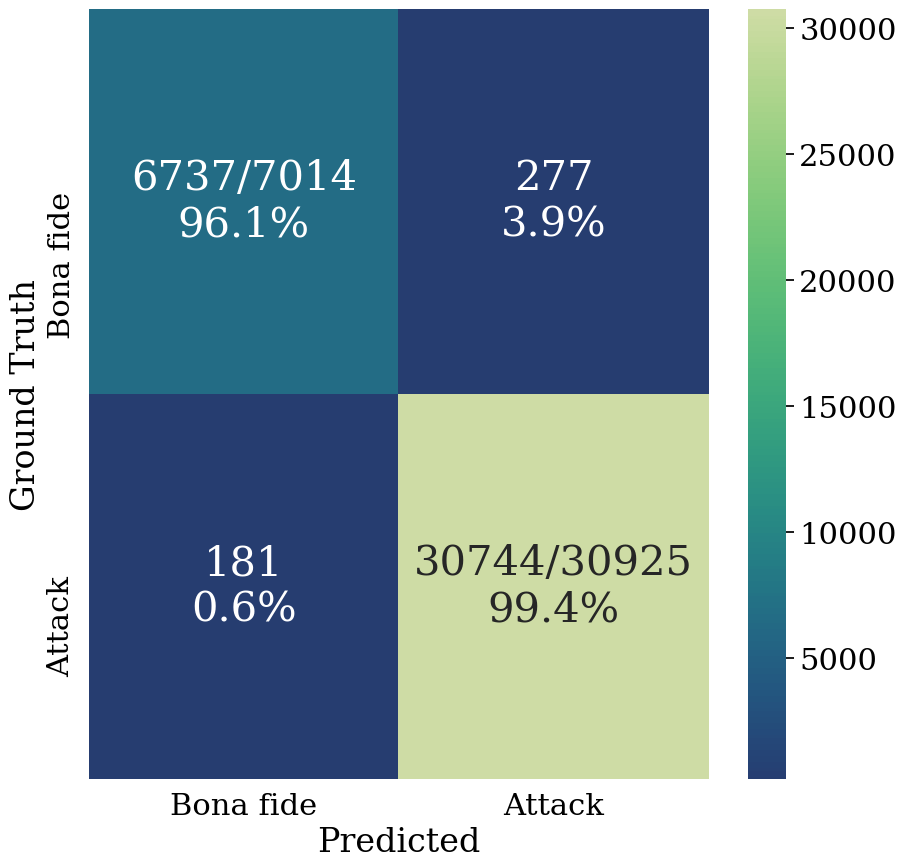}\label{fig:cascade1b}}\\
    \caption{Cascade models confusion matrices (a and b)}
    \label{fig:cascade1}
\end{figure*}

Figure~\ref{fig:cascade1a} shows the results reached for the concatenated system as a five-class problem. Figure~\ref{fig:cascade1b} show the confusion matrix for a binary problem, considering Composite, Print, Display, and Synthetic as one attack class. The most challenging scenario identified is the Display attack species. The Print species has a higher net amount of missclasified images, but most of these are classified as either Composite or Synthetic classes, which does not directly affect the final classification results since both of these classes are attack presentation species. See Figure~\ref{fig:cascade1b}.

Table~\ref{tab:cascade} shows the results for our two-stage proposal. This means a single image pass for both detectors is analysed. This approach is aligned with the purpose of the operational goal of PAD algorithms to correctly detect bona fide presentations (and not contribute to the system’s False Rejection Rate) while capturing as many attacks as possible. The highest PAIS values were obtained for the display attacks. This is due to the variability of screen resolutions (number of pixels per area), refresh ratio, post-processing, etc. applied per hardware to the picture to improve the visualisation, especially on smartphones. The two-stage network reached a combined BPCER\textsubscript{100} of 3.94\%.

\begin{table}[!htb]\def\tabularxcolumn#1{m{#1}}
    \centering
	\caption{Experiment 3 - Cascade models PAD assessment. The PAIS with the highest APCER in all cases is the \textbf{Display} class.}
	\label{tab:cascade}
	\begin{tabularx}{\linewidth}{lYYYY}
    	\toprule
    	\textbf{Metric} & \textbf{Border threshold} ($\tau$) & \textbf{Source threshold} ($\tau$) & \textbf{Combined APCER (\%)} & \textbf{Combined BPCER (\%)} \\
        \midrule
        EER                               &         0.9523  &         0.1054  &         1.4932  &         2.7802  \\
        BPCER\textsubscript{10}           &         0.0000  &         0.0002  &         9.9706  &         0.7984  \\
        BPCER\textsubscript{20}           &         0.0030  &         0.0033  &         4.8442  &         0.9980  \\
        BPCER\textsubscript{50}           &         0.5649  &         0.0506  &         1.9400  &         1.7821  \\
        \textbf{BPCER\textsubscript{100}} & \textbf{0.9844} & \textbf{0.2755} & \textbf{0.9289} & \textbf{3.9492} \\
    	\bottomrule
	\end{tabularx}
\end{table}

\section{Conclusion}
\label{sec:conclusion}

This work is one of the first end-to-end ID card PAD systems developed with more than 190,000 Chilean ID card images and several attack species using real images from an system on production. This new proposal outperforms by far our previous work based on 27,000 images and only three attacks species: Composite, Print, and Display. On the other hand, PyPAD has delivered new and complementary tools to plot DET curves, KDE and EER figures, and reporting of several ISO-compliant metrics.
We show that developing a two-stage system to detect attack presentations is more efficient and robust than a single multi class system. For future systems, we suggest dividing the attack per categories and train small classifiers. This fact allows us to improve and identify specific challenging PAIS. Selecting a proper image size for the input images, and avoiding image compression, post-processing and noise artifacts are critial to the operation of the classifiers. The minimum size of the image is important because images with a lower resolution than $448 \times448$ demonstrated sub-optimal results according to our experiments.

In summary, this work shows that developing and maintaining a PAD method for verification systems based on government-issued documents is critical due to the arms race nature of presentation attack detection research, given that these systems protect sensitive user information and are desirable targets for impostors. It is essential to consider an end-to-end system for a close-to-reality evaluation, including the image capture process, segmentation, and others. As future work, a country-agnostic PAD method capable of generalising documents for several different countries and document formats is needed.

\section*{Acknowledgments}

This research work has been partially funded by the German Federal Ministry of Education and Research and the Hessian Ministry of Higher Education, Research, Science and the Arts within their joint support of the National Research Center for Applied Cybersecurity ATHENE and TOC Biometrics R\&D Center.

\bibliographystyle{IEEEtran}
\bibliography{sample.bib}

\begin{thebibliography}{10}
\providecommand{\url}[1]{#1}
\csname url@samestyle\endcsname
\providecommand{\newblock}{\relax}
\providecommand{\bibinfo}[2]{#2}
\providecommand{\BIBentrySTDinterwordspacing}{\spaceskip=0pt\relax}
\providecommand{\BIBentryALTinterwordstretchfactor}{4}
\providecommand{\BIBentryALTinterwordspacing}{\spaceskip=\fontdimen2\font plus
\BIBentryALTinterwordstretchfactor\fontdimen3\font minus
  \fontdimen4\font\relax}
\providecommand{\BIBforeignlanguage}[2]{{%
\expandafter\ifx\csname l@#1\endcsname\relax
\typeout{** WARNING: IEEEtran.bst: No hyphenation pattern has been}%
\typeout{** loaded for the language `#1'. Using the pattern for}%
\typeout{** the default language instead.}%
\else
\language=\csname l@#1\endcsname
\fi
#2}}
\providecommand{\BIBdecl}{\relax}
\BIBdecl

\bibitem{iso30107}
{ISO/IEC JTC 1/SC 37 Biometrics}, ``{ISO/IEC} 30107-3, information technology
  --- biometric presentation attack detection --- part 3: Testing and
  reporting,'' International Organization for Standardization, Geneva, CH,
  Standard, 2021.

\bibitem{gonzalez2020hybrid}
S.~González, A.~Valenzuela, and J.~Tapia, ``Hybrid two-stage architecture for
  tampering detection of chipless {ID} cards,'' \emph{IEEE Transactions on
  Biometrics, Behavior, and Identity Science}, vol.~3, no.~1, pp. 89--100,
  2021.

\bibitem{imagenet_cvpr09}
J.~Deng, W.~Dong, R.~Socher, L.-J. Li, K.~Li, and L.~Fei-Fei, ``{ImageNet: A
  Large-Scale Hierarchical Image Database},'' in \emph{CVPR09}, 2009.

\bibitem{shi2018docface}
Y.~Shi and A.~K. Jain, ``Docface: Matching id document photos to selfies,'' in
  \emph{2018 IEEE 9th International Conference on Biometrics Theory,
  Applications and Systems (BTAS)}.\hskip 1em plus 0.5em minus 0.4em\relax
  IEEE, 2018, pp. 1--8.

\bibitem{shi2019docface+}
------, ``Docface+: Id document to selfie matching,'' \emph{IEEE Transactions
  on Biometrics, Behavior, and Identity Science}, vol.~1, no.~1, pp. 56--67,
  2019.

\bibitem{Zheng2019ASO}
L.~Zheng, Y.~Zhang, and V.~L.~L. Thing, ``A survey on image tampering and its
  detection in real-world photos,'' \emph{J. Vis. Commun. Image Represent.},
  vol.~58, pp. 380--399, 2019.

\bibitem{albiero}
V.~Albiero, N.~Srinivas, E.~Villalobos, J.~Perez-Facuse, R.~Rosenthal, D.~Mery,
  K.~Ricanek, and K.~W. Bowyer, ``Identity document to selfie face matching
  across adolescence,'' in \emph{2020 IEEE International Joint Conference on
  Biometrics (IJCB)}, 2020, pp. 1--9.

\bibitem{Bulan}
O.~Bulan and G.~Sharma, ``Content authentication for printed images utilizing
  high capacity data hiding,'' \emph{Journal of Electronic Imaging}, vol.~22,
  p. 033006, 07 2013.

\bibitem{stokkenes2018biometric}
M.~Stokkenes, R.~Ramachandra, and C.~Busch, ``{Biometric Transaction
  Authentication using Smartphones},'' in \emph{2018 International Conference
  of the Biometrics Special Interest Group (BIOSIG)}.\hskip 1em plus 0.5em
  minus 0.4em\relax IEEE, 9 2018, pp. 1--5.

\bibitem{perera2019face}
P.~Perera and V.~M. Patel, ``{Face-Based Multiple User Active Authentication on
  Mobile Devices},'' \emph{IEEE Transactions on Information Forensics and
  Security}, vol.~14, no.~5, pp. 1240--1250, 5 2019.

\bibitem{fathy2015face}
M.~E. Fathy, V.~M. Patel, and R.~Chellappa, ``{Face-based Active Authentication
  on Mobile Devices},'' in \emph{2015 IEEE International Conference on
  Acoustics, Speech and Signal Processing (ICASSP)}.\hskip 1em plus 0.5em minus
  0.4em\relax IEEE, 4 2015, pp. 1687--1691.

\bibitem{MIDV-500}
V.~V. Arlazarov, K.~B. Bulatov, and T.~S. Chernov, ``{MIDV-500:} {A} dataset
  for identity documents analysis and recognition on mobile devices in video
  stream,'' \emph{CoRR}, vol. abs/1807.05786, 2018.

\bibitem{zhu}
X.~Zhu, H.~Liu, Z.~Lei, H.~Shi, F.~Yang, D.~Yi, G.~Qi, and S.~Z. Li,
  ``Large-scale bisample learning on {ID} versus spot face recognition,''
  \emph{Int. J. Comput. Vis.}, vol. 127, no. 6-7, pp. 684--700, 2019.

\bibitem{DBLP:journals/iet-bmt/RajaR022}
R.~Mudgalgundurao, P.~Schuch, K.~B. Raja, R.~Raghavendra, and N.~Damer,
  ``Pixel-wise supervision for presentation attack detection on id cards,''
  \emph{{IET} Biometrics}, 2022.

\bibitem{brummer2013bosaris}
N.~Brümmer and E.~de~Villiers, ``The {BOSARIS} toolkit: Theory, algorithms and
  code for surviving the new {DCF},'' \emph{arXiv preprint arXiv:1304.2865},
  2013.

\bibitem{larcher2016extensible}
A.~Larcher, K.~A. Lee, and S.~Meignier, ``An extensible speaker identification
  sidekit in python,'' in \emph{2016 IEEE International Conference on
  Acoustics, Speech and Signal Processing (ICASSP)}.\hskip 1em plus 0.5em minus
  0.4em\relax IEEE, 2016, pp. 5095--5099.

\bibitem{zhang2016joint}
K.~Zhang, Z.~Zhang, Z.~Li, and Y.~Qiao, ``Joint face detection and alignment
  using multitask cascaded convolutional networks,'' \emph{IEEE Signal
  Processing Letters}, vol.~23, no.~10, pp. 1499--1503, 2016.

\bibitem{danben_synt}
\BIBentryALTinterwordspacing
D.~Benalcazar, J.~E. Tapia, S.~Gonzalez, and C.~Busch, ``Synthetic id card
  image generation for improving presentation attack detection,'' 2022.
  [Online]. Available: \url{https://arxiv.org/abs/2211.00098}
\BIBentrySTDinterwordspacing

\bibitem{ronneberger2015unet}
O.~Ronneberger, P.~Fischer, and T.~Brox, ``{U-Net}: Convolutional networks for
  biomedical image segmentation,'' in \emph{International Conference on Medical
  Image Computing and Computer-Assisted Intervention -- MICCAI 2015}, N.~Navab,
  J.~Hornegger, W.~M. Wells, and A.~F. Frangi, Eds.\hskip 1em plus 0.5em minus
  0.4em\relax Springer, 2015, pp. 234--241.

\bibitem{lara2021towards}
R.~Lara, A.~Valenzuela, D.~Schulz, J.~Tapia, and C.~Busch, ``Towards an
  efficient semantic segmentation method of {ID} cards for verification
  systems,'' \emph{arXiv preprint arXiv:2111.12764}, 2021.

\bibitem{marcel2019handbook}
S.~Marcel, M.~S. Nixon, J.~Fierrez, and N.~W.~D. Evans, Eds., \emph{Handbook of
  Biometric Anti-Spoofing - Presentation Attack Detection, Second Edition},
  ser. Advances in Computer Vision and Pattern Recognition.\hskip 1em plus
  0.5em minus 0.4em\relax Springer, 2019.

\end{thebibliography}
\vspace{-0.3cm}

\begin{IEEEbiography}[{\includegraphics[width=1in,height=1.25in,clip,keepaspectratio]{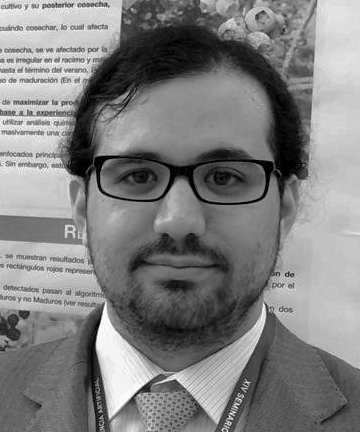}}]{Sebastian Gonzalez} received a B.Sc. in Computer Science from Universidad Andres Bello in 2019. He is currently researcher at TOC Biometrics. Currently, he is pursuing a M.Sc. degree at Universidad de Santiago de Chile. His main interests include topics such as presentation attack detection, classification, segmentation, and pattern recognition.
\end{IEEEbiography}
\vspace{-0.3cm}

\begin{IEEEbiography}
[{\includegraphics[width=1in,height=1.25in,clip,keepaspectratio]{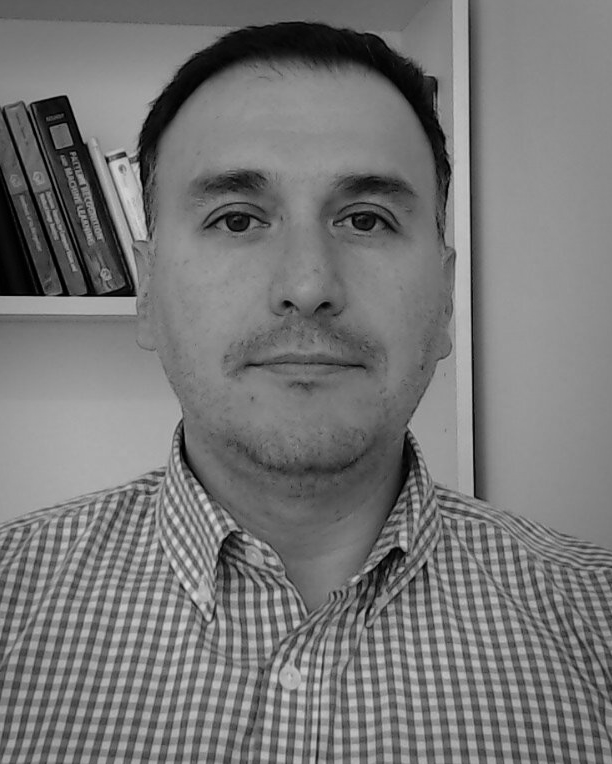}}]{Juan Tapia} received a P.E. degree in Electronics Engineering from Universidad Mayor in 2004, a M.S. in Electrical Engineering from Universidad de Chile in 2012, and a Ph.D. from the Department of Electrical Engineering, Universidad de Chile in 2016. In addition, he spent one year of internship at the University of Notre Dame. In 2016, he received the award for best Ph.D. thesis. From 2016 to 2017, he was an Assistant Professor at Universidad Andres Bello. From 2018 to 2020, he was the R\&D Director for the area of Electricity and Electronics at Universidad Tecnologica de Chile and R\&D Director of TOC Biometrics from 2019 up to 2022. Currently, he is a Senior Researcher at Hochschule Darmstadt (HDA). His main research include pattern recognition and deep learning applied to iris biometrics, morphing, feature fusion, and feature selection. 
\end{IEEEbiography}


\vfill

\end{document}